\documentclass[10pt,twocolumn,letterpaper]{article}

\pdfoutput=1

\usepackage{iccv}
\usepackage{times}
\usepackage{epsfig}
\usepackage{graphicx}
\usepackage{amsmath}
\usepackage{amssymb}
\usepackage{subcaption}
\usepackage{enumitem}
\usepackage{multirow}
\usepackage{multicol}
\usepackage{float}
\newcommand{\authorblock}[1]{\begin{tabular}{@{}c@{}}#1\end{tabular}}

\usepackage[pagebackref=true,breaklinks=true,letterpaper=true,colorlinks,bookmarks=false]{hyperref}

\iccvfinalcopy 

\def\iccvPaperID{6480} 

\ificcvfinal\fi
\begin{document}

\title{Single-Network Whole-Body Pose Estimation}

\author{\begin{tabular}{ccc}
  \authorblock{{\normalsize Gines Hidalgo$^1$, Yaadhav Raaj$^1$, Haroon Idrees$^2$, Donglai Xiang$^1$, Hanbyul Joo$^3$, Tomas Simon$^1$, Yaser Sheikh$^1$}} \\
  \authorblock{{\tt\normalsize $^1$Carnegie Mellon University, $^2$RetailNext, $^3$Facebook AI Research }} \\
  \authorblock{{\normalsize
    gines.hidalgo@epicgames.com,
    \{ryaadhav,donglaix,yaser\}@cs.cmu.edu,
    haroon@retailnext.net,
    \{hjoo,tsimon\}@fb.com
  }} \\
\end{tabular}\vspace{.09in}}


\maketitle

\begin{abstract}\vspace{-.1in}
We present the first single-network approach for 2D~whole-body pose estimation, which entails simultaneous localization of body, face, hands, and feet keypoints. Due to the bottom-up formulation, our method maintains constant real-time performance regardless of the number of people in the image. The network is trained in a single stage using multi-task learning, through an improved architecture which can handle scale differences between body/foot and face/hand keypoints. Our approach considerably improves upon OpenPose~\cite{cao2018openpose}, the only work so far capable of whole-body pose estimation, both in terms of speed and global accuracy. Unlike \cite{cao2018openpose}, our method does not need to run an additional network for each hand and face candidate, making it substantially faster for multi-person scenarios. This work directly results in a reduction of computational complexity for applications that require 2D whole-body information (e.g., VR/AR, re-targeting). In addition, it yields higher accuracy, especially for occluded, blurry, and low resolution faces and hands. For code, trained models, and validation benchmarks, visit our project page\footnote{\url{https://github.com/CMU-Perceptual-Computing-Lab/openpose_train}}.

\end{abstract}


\section{Introduction}
Human keypoint estimation has been an open problem for decades in the research community. Initially, efforts were focused on facial alignment (i.e., face keypoint detection)~\cite{asthana2014incremental, ranjan2019hyperface, saragih2009face, zhang2016joint, xiong2013supervised, zhang2016learning}, and later evolved into single and multi-person human pose estimation in-the-wild, including body and foot keypoints~\cite{Andriluka2010, belagiannis2017recurrent, cao2017realtime, chen2017cascaded, Dantone2013}. A more recent and challenging problem has targeted hand keypoint detection~\cite{simon2017hand, sridhar2013interactive, zimmermann2017learning}. Therefore, the next logical step is the integration of all of these keypoint detection tasks within the same algorithm, leading to ``whole-body" or ``full-body" (body, face, hand, and foot) pose estimation~\cite{alp2018densepose, cao2018openpose}.


There are several applications that can immediately take advantage of whole-body keypoint detection, including augmented reality, virtual reality, medical applications, and sports analytics. Whole-body keypoint detection can also provide more subtle cues for re-targeting, 3D human keypoint and mesh reconstruction~\cite{Recycle-GAN, chan2018everybody, joo2018total, panteleris2017using, xiang2018monocular}, person re-identification, tracking, and action recognition
~\cite{gui2018teaching, mehta2017vnect, raaj2018efficient, qian2017pose}. Despite these needs, the only existing method providing whole-body pose estimation is the prior version of OpenPose~\cite{cao2018openpose}, which follows a multi-stage approach. First, all body poses are obtained from an input image in a bottom-up fashion~\cite{cao2017realtime} and then additional face and hand keypoint detectors are run for each detected person~\cite{simon2017hand}.
As a multi-network approach, it directly uses the existing body, face, and hand keypoint detection algorithms. However, it suffers from the issue of early commitment: there is no recourse to recovery if the body-only detector fails, especially during partial visibility when only the face or hand are visible in the image. In addition, its run-time is proportional to the number of people in the image, making whole-body pose estimation prohibitively expensive for multi-person and real-time applications. A single-stage method, estimating whole-body poses of multiple people in a single pass, would be more attractive as it would yield a fixed inference run-time, independent of the number of people in the scene. 

Unfortunately, there is an inherent scale difference between body/foot and face/hand keypoints.
The former require a large receptive field to learn the complex interactions across people (contact, occlusion, limb articulation), while the latter require 
higher image resolution. Since the foot pose is highly dependent on that of the body, unlike face and hands, its desirable scale is consistent with that of the body. Moreover, the scale issue has two critical consequences. First, datasets with full-body annotations in-the-wild do not currently exist, since the characteristics of each set of keypoints result in different kinds of datasets. Body datasets predominantly contain images with multiple people, usually resulting in fairly low face and hand resolution, while face and hand datasets mostly contain images with a single, cropped face or hand. Secondly, the architecture design of a single-network model must differ from that of the state-of-the-art keypoint detectors in order to offer high-resolution and a larger receptive field, while simultaneously improving the inference run-time of multi-network approaches.

To overcome the dataset problem, we resort to multi-task learning (MTL), a classic machine learning technique~\cite{girshick2015fast, misra2016cross, zhang2014facial} where related learning tasks are solved simultaneously by exploiting commonalities and differences across them. Previously, MTL has been successful in training a combined body-foot keypoint detector~\cite{cao2018openpose}. Nevertheless, it does not generalize to whole-body estimation because of the underlying scale problem. Therefore, the major contributions of this paper are summarized as follows:
\begin{itemize}[noitemsep]
    \item \textbf{Novelty:} We present an MTL approach combined with an improved architecture design to train a unified model for various keypoint detection tasks each with different scale characteristics. 
    This results in the \textit{first} single-network approach for whole-body multi-person pose estimation. 
    \item \textbf{Speed:}
    At test time, our single-network approach provides a constant real-time inference regardless of the number of people detected, and it is approximately $n$ times faster than the state-of-the-art (OpenPose~\cite{cao2018openpose}) for images with $n$ people. In addition, it is trained in a single stage, rather than requiring independent network training for each individual task. This reduces the total training time approximately by one-half.
    \item \textbf{Accuracy:} Our approach also yields higher accuracy than that of the previous OpenPose, especially for face and hand keypoint detection, generalizing better to occluded, blurry, and low resolution faces and hands.
\end{itemize}

\section{Related Work}
\textbf{Face Keypoint Detection:} Also referred in literature as landmark detection or face alignment, it has a long history in computer vision and many approaches have been proposed to tackle it. These approaches are broadly divisible into two categories:
template fitting~\cite{asthana2014incremental, kazemi2014one, saragih2009face, xiong2013supervised, zhu2015face} and regression-based
methods~\cite{ranjan2019hyperface, zhang2016joint, zhang2016learning}.
Template fitting methods build face templates to fit input images, usually exploiting a cascade of regression functions.
Regression methods,
on the other hand,
are based on Convolutional Neural Networks (CNNs) and usually apply convolutional heatmap regression. They operate in a similar fashion to that of body pose estimation.

\textbf{Body Keypoint Estimation:}
With the face alignment problem solved, efforts moved towards single-person pose estimation. The initial approaches performed inference over both local observations on body parts and their spatial dependencies, either based on tree-structured graphical models~\cite{Andriluka2010,fh2005pictorial,pishchulin2013poselet,ramanan2005strike,yang2013articulated} 
or non-tree models~\cite{Dantone2013,karlinsky2012using,lan2005beyond,sigal2006measure,wang2008multiple}.
The popularity of CNNs and the release of massive annotated datasets (COCO~\cite{lin2014microsoft} and MPII~\cite{andriluka20142d})
imparted significant boost in the accuracy
of single-person estimation~\cite{belagiannis2017recurrent,chen2017adversarial,chu2017multi,Ke_2018_ECCV,newell2016stacked,tang2018deeply,wei2016cpm,yang2017learning}, 
and 
have enabled multi-person estimation. The latter is traditionally divided into top-down~\cite{chen2017cascaded,fang2017rmpe,gkioxari2014using,he2017mask,iqbal2016multi,papandreou2017towards,pishchulin2012articulated,xiao2018simple} 
and bottom-up~\cite{cao2017realtime, newell2017associative, nie2018ppn, papandreou2018personlab, pishchulin2015deepcut} 
approaches.

\textbf{Foot Keypoint Estimation:}
Cao~\etal~\cite{cao2018openpose} released the first foot dataset, with annotations on a subset of images from the COCO dataset.
They also trained the first combined body-foot keypoint detector
by applying a naive multi-task learning technique. 
Our method is an extension of this work, mitigating its limitations and enabling it to generalize to both large-scale body and foot keypoints as well as the more subtle face and hand keypoints. 

\begin{figure*}[t]
    \centering
    \includegraphics[width=1\linewidth]{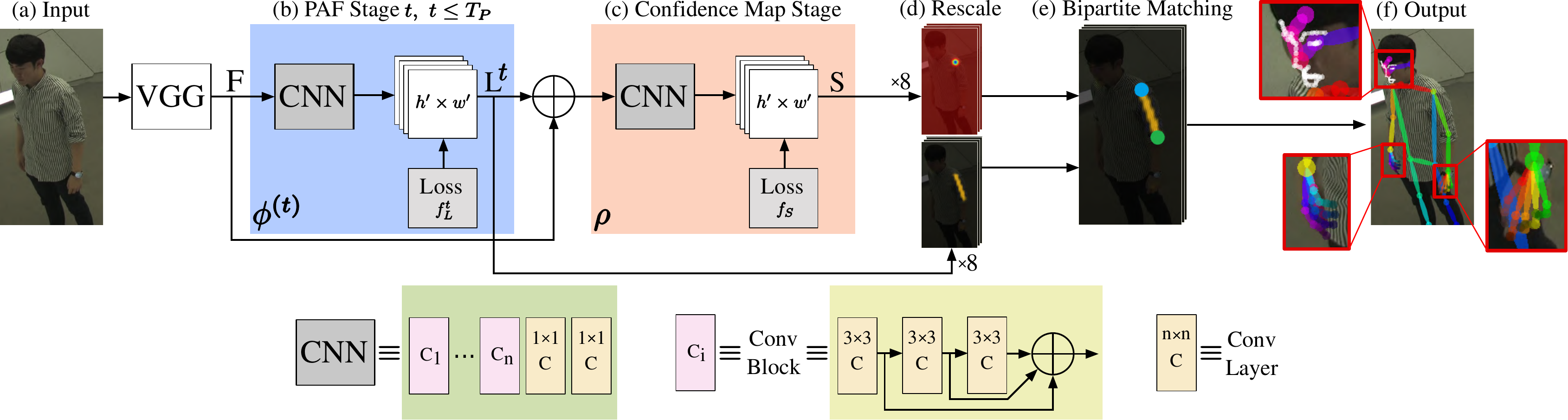}
    \caption{Overall pipeline.
    (a)~An RGB image is taken as input.
    (b,c)~Architecture of the 
    whole-body pose estimation network, consisting of multiple stages predicting refined PAFs ($\mathbf{L}$) and confidence maps ($\mathbf{S}$) for body, face, hand and foot.
    It is trained end-to-end with a multi-task loss that combines the losses of each individual keypoint annotation task.    
    Each \textit{Conv Layer}, $\mathrm{C}$,  corresponds to a Convolution-PReLU sequence.
   (d)~At test time, the most refined PAFs and confidence maps are resized to increase the accuracy. 
   (e)~The parsing algorithm uses the PAFs to find all the whole-body parts belonging to the same person using bipartite matching.
   (f)~The final whole-body poses are returned for all the people in the image.
   }
   \label{fig:architecture}
\end{figure*}


\textbf{Hand Keypoint Detection:}
With the exciting improvements in face and 
body estimation, recent research is targeting hand keypoint detection. However, its manual annotation is extremely challenging and expensive due to heavy self-occlusion~\cite{simon2017hand}. As a result, large hand keypoint datasets do not exist in-the-wild. To alleviate this problem, early work 
is based on depth information~\cite{oikonomidis2012tracking, sharp2015accurate, sridhar2015fast, sridhar2013interactive}, 
but is limited to indoor scenarios. 
Most of the work in RGB-based hand estimation is focused on 3D estimation~\cite{cai2018weakly, iqbal2018hand, mueller2018ganerated, zimmermann2017learning}, primarily based on fitting complex 3D models with strong priors. In the 2D RGB domain, Simon \textit{et al.}~\cite{simon2017hand} exploit multi-view bootstrapping to create a hand keypoint dataset and train a 2D RGB-based hand detector. First, a naive detector is trained on a small subset of manually labeled annotations.
Next, this detector is applied in a 30-camera multi-view dome structure~\cite{Joo_2015_ICCV, joo2018total} to obtain new annotations based on 3D reconstruction.
Unfortunately, most of the methods have only demonstrated results in controlled lab environments. 

\textbf{Whole-Body Keypoint Detection:}
OpenPose~\cite{cao2018openpose, cao2017realtime, simon2017hand} is the only known work able to provide all body, face, hand, and foot keypoints in 2D. It operates in a multi-network fashion. First, it detects the body and foot keypoints based on~\cite{cao2017realtime, wei2016cpm}. Then, it approximates the face and hand bounding boxes based on the body keypoints, and applies a keypoint detection network for each subsequent face and hand candidate~\cite{simon2017hand}. Recent work is also targeting 3D mesh reconstruction~\cite{joo2018total, kanazawa2018end}, usually leveraging the lack of 3D datasets with the existing 2D datasets and detectors, or reconstructing the 3D surface of the human body from denser 2D human annotations~\cite{alp2018densepose}.

\textbf{Multi-Task Learning:}
To overcome the problems of state-of-the-art whole-body pose estimation, we aim to apply multi-task learning (MTL) to train a single whole-body estimation model out of the four different tasks: body, face, hand, and foot detection. MTL applied to deep learning can be split into soft and hard parameter sharing of hidden layers. In soft parameter sharing, each task has its own model, but the distance between the parameters is regularized to encourage them to be similar between models~\cite{duong2015low, yang2016trace}. Hard parameter sharing is the most commonly used MTL approach in computer vision, applied in many applications, such as facial alignment~\cite{zhang2014facial} or surface normal prediction~\cite{misra2016cross}. Particularly, it has had a critical impact on object detection, where popular approaches such as Fast R-CNN~\cite{girshick2015fast} exploit MTL to merge all previously independent object detection tasks into a single and improved detector. It considerably improved training and testing speeds as well as detection accuracy. Analogous to Fast R-CNN, our work brings together multiple and independent keypoint detection tasks into a unified framework. See~\cite{ruder2017overview} for a more detailed survey of multi-task learning literature.

\section{Method}\label{sec:method}
Our system follows a streamlined approach, using an RGB image to generate a set of whole-body human keypoints for each person detected. This global pipeline is illustrated in Fig.~\ref{fig:architecture}. The extracted keypoints contain information from the face, torso, arms, hands, legs, and feet.
The network architecture of the proposed whole-body keypoint detector can be based on any state-of-the-art body-only keypoint detector.
For a fair comparison of our results with previous versions of OpenPose~\cite{cao2018openpose, cao2017realtime, simon2017hand}, we reuse its Part Affinity Field (PAF) network architecture.

\subsection{PAF-based Body Pose Estimation}
Here, we review the main details of the PAF-based method. We refer the reader to~\cite{cao2017realtime} for a full description. 
This approach iteratively predicts Part Affinity Fields (PAFs), which encode part-to-part associations, and detection confidence maps. Each PAF is defined as a 2D orientation vector that points from one keypoint to another. 
The input image $I$ is initially analyzed by a convolutional network (pre-trained on VGG-19~\cite{simonyan2014very}), generating a set of feature maps $\mathbf{F}$. Next, $\mathbf{F}$ is fed into the first stage $\phi^{(1)}$ of the network $\phi$, which predicts a set of PAFs $\mathbf{L}^{(1)}$. For each subsequent stage $i$, the PAFs of the previous stage $\mathbf{L}^{(t-1)}$ are concatenated to $\mathbf{F}$ and refined to produce $\mathbf{L}^{(t)}$. After $N$ stages, we obtain the final set of PAF channels $\mathbf{L} = \mathbf{L}^{(N)}$. Then, $\mathbf{F}$ and $\mathbf{L}$ are concatenated and fed into a network $\rho$, which predicts the keypoint confidence maps $\mathbf{S}$, i.e.,
\begin{gather}
\mathbf{L}^{(1)}=\phi^{(1)}\left(\mathbf{F}\right),
\\
\mathbf{L}^{(t)}=\phi^{(t)}\left(\mathbf{F},\mathbf{L}^{(t-1)}\right),\:\text{\ensuremath{\forall}}~2\leq t\leq N,
\\
\mathbf{L}=\mathbf{L}^{(N)},
\\
\mathbf{S}=\rho\left(\mathbf{F},\mathbf{L}\right).
\end{gather}

An $\ell$2 loss function is applied at the end of each stage, which compares the estimated predictions and the groundtruth maps ($\mathbf S^*$) and fields ($\mathbf L^*$) for each pixel ($p$) on each confidence map ($c$) and PAF ($f$) channel:
\begin{gather}
f_{\mathbf{L}}=\sum_{f=1}^F\sum_{p}\left(W_i(p) \cdot \Vert\mathbf{L}_f(p)-\mathbf{L}_f^*(p)\Vert_2^2\right),
\\
f_{\mathbf{S}}=\sum_{c=1}^{C}\sum_{p}\left(W_i(p) \cdot \Vert\mathbf{S}_c(p)-\mathbf{S}_c^*(p)\Vert_2^2\right),
\end{gather}
where $C$ and $F$ are the number of stages for confidence map and PAF prediction, and $W$ is a binary mask with $W_i(p)$=0 when an annotation is missing at a pixel 
$p$ for a particular confidence map or PAF channel $i$. Non-maximum suppression is performed on the confidence maps to obtain a discrete set of body part candidate locations. Finally, bipartite graph matching~\cite{west1996introduction} is used to assemble the connections that share the same part detection candidates into full-body poses for each person in the image.

\subsection{Whole-Body Pose Estimation}
We want whole-body pose estimation to be accurate but also fast. Training an individual PAF-based network to predict each individual set of keypoints would achieve the first goal, but would also be computationally inefficient. Instead, we extend the body-only PAF framework to whole-body pose estimation, making various modifications to the training approach and network architecture.

\textbf{Multi-task learning training:}
We modify the definition of the keypoint confidence maps $\mathbf{S}$ as the concatenation of the confidence maps: body $S_B$, face $S_F$, hand $S_H$, and foot $S_O$. Analogously, the set of PAFs at stage $t$, $\mathbf{L}^{(t)}$, is defined as the concatenation of the PAFs: body $\mathbf{L}_B^{(t)}$, face $\mathbf{L}_F^{(t)}$, hand $\mathbf{L}_H^{(t)}$, and foot $\mathbf{L}_O^{(t)}$.
An interconnection between the different annotation tasks must 
be created in order to allow the different
set of
keypoints of the same person to be assembled together. For instance, we join the body and foot keypoints through the ankle keypoint, which is annotated in both datasets. Analogously, the wrists connect the body and hand keypoints, while the eyes relate body and face.
The rest of the pipeline (non-maximum suppression over confidence maps and bipartite matching to assemble full people) is left intact.
As opposed to having a dedicated network for each keypoint annotation task, all the keypoints are now defined within the same model architecture. This is an extreme version of hard parameter sharing, in which only the final layer is task-specific.

\textbf{Balanced dataset-based probability ratio:}
If we had a whole-body dataset, we could train a combined model following the body-only training approach. Unfortunately, each available dataset only contains annotations for a subset of keypoints.
To overcome the lack of a combined dataset, we follow the probability ratio idea of the single-network body-foot detector of Cao~\etal~\cite{cao2018openpose}, which was trained from body-only and body-foot datasets.
Batches
of images are randomly picked from each available dataset, and the losses for the confidence map and PAF channels associated to non-labeled keypoints are masked out. I.e., their binary mask $W_i(p)$ is set to 0. Abusing notation, the probability ratio $P^d$ is defined as the probability of picking the next annotated batch of images from the dataset $d$. This probability is distributed across the different datasets depending on the number of images in each dataset. 
When applied to keypoints with similar scale properties (e.g., body and foot~\cite{cao2018openpose}), it results in a robust keypoint detector. However,  it does not converge when applied to whole-body estimation, which now includes face and hand keypoints. Additionally, the accuracy of the body and foot detectors is considerably reduced.
Solving the face and hand convergence problem would be then possible through
a deeper analysis of the properties and differences of each set of keypoints.

\begin{figure}[t]
\centering
    \begin{subfigure}[t]{0.325\linewidth}
        \centering
        \includegraphics[width=1\linewidth]{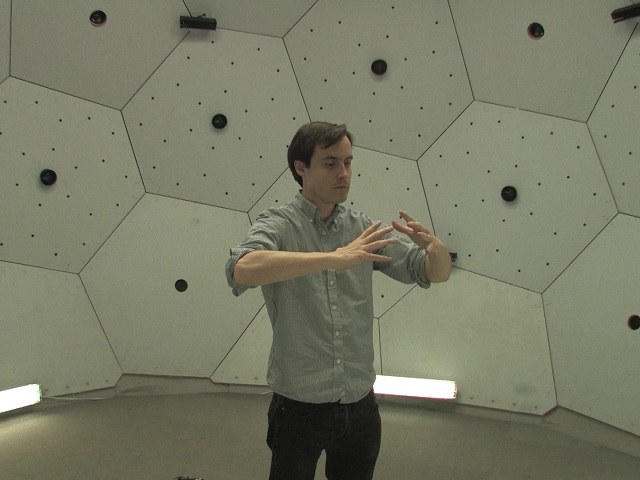}
        \caption{}
    \end{subfigure}
    \begin{subfigure}[t]{0.325\linewidth}
        \centering
        \includegraphics[width=1\linewidth]{figures/body.jpg}
        \caption{}
    \end{subfigure}
    \begin{subfigure}[t]{0.325\linewidth}
        \centering
        \includegraphics[width=1\linewidth]{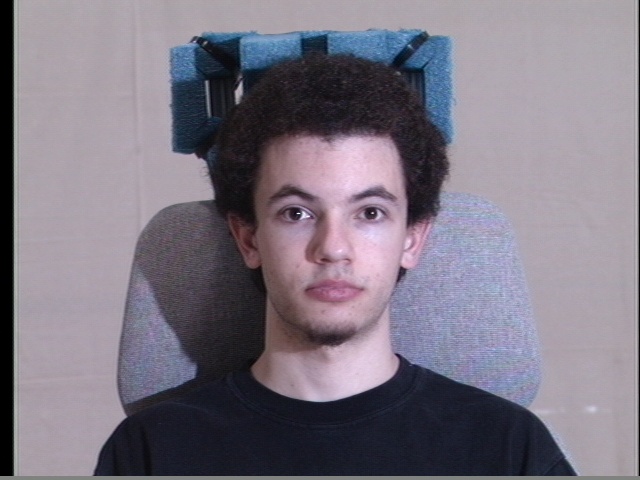}
        \caption{}
    \end{subfigure}
   \caption{Different kinds of datasets for each set of keypoints present different properties (number of people, occlusion, person scale, etc.).
   We show typical examples from the hand (left), body (center), and face (right) datasets.
   }
\label{fig:dataset_examples}
\end{figure}

\textbf{Dataset-based augmentation:}
There is an inherent scale difference between body/foot and face/hand keypoints,
which results in different kinds of datasets for each set of keypoints. Body datasets predominately contain images with multiple people and low face and hand resolution; face datasets focus on images with a single person or cropped face; whereas hand datasets usually contain images with a single full-body person. Fig.~\ref{fig:dataset_examples} shows typical examples from each dataset.
To solve this problem, augmentation parameters are varied for each set of keypoints. For instance, the minimum possible scale for face datasets is reduced during data augmentation to expose our model to small faces, 
to simulate reality of the `wild' environments. Oppositely, the maximum scale for hand datasets is increased so that full-sized hands appear more frequently, allowing the network to generalize to high resolution hands.

\textbf{Overfitting:}
Using the above ideas, the face and hand detectors converge and allow us to build an initial whole-body pose detector.
However, we observe a large degree of over-fitting on some validation sets, particularly in the face and lab-recorded datasets. Even though the initial probability ratio $P^d$ is evenly distributed depending on the number of images in each dataset, the data complexity of these datasets is lower than the complexity of the challenging multi-person and in-the-wild datasets. In addition, the range of possible facial gestures is much smaller than the number of possible body and hand poses. Thus, the probability ratio of picking a batch from one of the face and lab-recorded datasets must be additionally reduced. Empirically, we fine-tune the probability ratios between datasets so that the validation accuracy on each converges at the same pace.

\textbf{High false positive rate:}
Face, hand, and foot keypoints present a high false positive rate producing a ``ghosting" effect on their respective confidence map and PAF channels. Visually, this means that these channels are outputting a non-zero value in regions that do not contain people. To mitigate this issue,
their binary mask $W_i(p)$
is re-enabled in the COCO dataset for image regions with no people. Furthermore, we complement training with an additional dataset consisting of COCO images without any people.

\textbf{Further refinement:}
Face and hand datasets do not necessarily annotate all the people that appear in each image. 
We apply Mask R-CNN~\cite{he2017mask} to mask out the regions of the image with non-labeled people. 
In addition, the pixel localization precision of the face and hand keypoint detectors remains low. To moderately improve it, we reduce the radius of the Gaussian distribution used to generate the groundtruth of their confidence map channels. 

\textbf{Shallow whole-body detector:}
At this point, we can build a working whole-body pose detector.
The inference run-time of this refined detector matches that of running body-foot in~\cite{cao2018openpose}. However, it still suffers from two main issues. On the one hand, the body and foot accuracy considerably decreases compared to its standalone analog (i.e., the body-foot detector from Cao~\etal~\cite{cao2018openpose}). The complexity of the network output has increased from predicting 25 to 135 keypoints (and their corresponding PAFs). The network has to compress about 5 times more information with the same number of parameters, reducing the accuracy of each individual part. On the other hand, face and hand detection accuracy appear relatively similar to that of Cao~\etal~\cite{cao2018openpose} in the benchmarks, but the qualitative results show that their pixel localization precision remains low. This is due to the reuse of same network as that used in body-only pose estimation which has low input resolution. Face and hand detection requires a network with higher resolution to provide results with high pixel localization precision. This initial detector is defined as ``Shallow whole-body" in Sec.~\ref{sec:evaluation}.

\textbf{Improved network architecture:}
To match the accuracy of the body-only detector and solve the resolution issue of face and hand, the whole-body network architecture must diverge from that of Cao~\etal~\cite{cao2018openpose}. It must still maintain a large receptive field for accurate body detection but also offer high-resolution maps for precise face and hand keypoint detection. Additionally, its inference run-time should remain similar to or improve upon that of its analogous multi-stage whole-body detector. Our final model architecture, refined for whole-body estimation and shown in Fig.~\ref{fig:architecture}, differs from the original baseline in the following details:
\begin{itemize}[noitemsep]
    \item The network input resolution is increased to considerably improve face and hand precision. Unfortunately, this implicitly reduces the effective receptive field (further reducing body accuracy). 
    \item The number of convolutional blocks in each PAF stage is increased to recover the effective receptive field that was previously reduced. 
    \item The width of each convolutional layer in the last PAF stage is increased to improve the overall accuracy, enabling our model to match the body accuracy of the standalone body detector. 
    \item The previous solutions considerably increase the overall accuracy of our approach but also harm the training and testing speed. The number of PAF stages is reduced to partially overcome this issue, which only results in a moderate reduction in overall accuracy.
\end{itemize}

This improved model highly outperforms Cao~\etal~\cite{cao2018openpose} in speed, being approximately $n$ times faster for an image with $n$ people in it. Additionally, it also slightly improves its global accuracy (Secs.~\ref{sec:body_val}, \ref{sec:face_val} and \ref{sec:hand_val}).
This network is denoted as ``Deep whole-body" in Sec.~\ref{sec:evaluation}.

\section{Evaluation}\label{sec:evaluation}

\subsection{Experimental Setup}\label{sec:param_config}
\textbf{Datasets:}
We train and evaluate our method on different benchmarks for each set of keypoints:
(1)~COCO keypoint 
dataset~\cite{lin2014microsoft} for multi-person body estimation;
(2)~OpenPose foot dataset~\cite{cao2018openpose}, which is a subset of 15k annotations out of the COCO keypoint dataset;
(3)~OpenPose hand dataset~\cite{simon2017hand}, which combines a subset of 1k hand instances manually annotated from MPII~\cite{andriluka20142d} as well as a set of 15k samples automatically annotated on the Dome or Panoptic Studio~\cite{Joo_2017_TPAMI};
(4)~our custom face dataset, consisting of a combination of the CMU Multi-PIE Face~\cite{gross2010multi}, Face Recognition Grand Challenge (FRGC)~\cite{frgc_2010}, and i-bug~\cite{sagonas2016300} datasets;
(5)~the Monocular Total Capture dataset~\cite{xiang2018monocular}, the only available 2D whole-body dataset which has been recorded in the same Panoptic Studio used for the hand dataset. 
Following the standard COCO multi-person metrics, we report mean Average Precision (AP) and mean Average Recall (AR) for all sets of keypoints. 

\textbf{Training:} All models are trained using 4-GPU machines, with a batch size of 10 images, Adam optimization, and an initial learning rate of 5e-5. We also decrease the learning rate by a factor of 2 after 200k, 300k, and every additional 60k iterations. We apply random cropping, rotation ($\pm$45$^o$), flipping (50\%), and scale (in the range $[1/3, 1.5]$) augmentation. The scale is modified to $[2/3, 4.5]$ and $[0.5, 4.0]$ for Dome and MPII hand datasets, respectively. The input resolution of the network is set to $480$$\times$$480$ pixels. Similar to~\cite{cao2018openpose}, we maintain VGG-19 as the backbone. The probability of picking an image from each dataset is 76.5\% for COCO, 5\% each for foot and MPII datasets, 0.33\% for each face dataset, 0.5\% for Dome hand, 5\% for MPII hand, 5\% for whole-body data, and 2\% for picking an image with no people in it.

\textbf{Evaluation:} We report both single-scale (image resized to a height of 480 pixels while maintaining the aspect ratio) and multi-scale results
(results averaged from images resized to a height of 960, 720, 480, and 240 pixels).



\subsection{Ablation Experiments}
Increasing the network resolution is crucial for accurate hand and face detection. Nevertheless, it directly results in slower training and testing speeds. We aim to maximize the accuracy while preserving
a reasonable run-time performance.
Thus, we explore multiple models tuned to maintain the same inference run-time. The final model is selected as the one maximizing the body AP.
Table~\ref{table:self_coco_val} shows the results on the COCO~\cite{lin2014microsoft} validation set. 
The most efficient configuration is achieved by increasing the number of convolutional blocks and their width, while reducing the number of stages in order to preserve the speed. 

\begin{table}[h]
    \centering
    \resizebox{\columnwidth}{!}{
    \begin{tabular}{l|l|c|c|c|c}
    \hline
    \multicolumn{2}{c|}{Method}    & \multirow{2}{*}{AP} & \multirow{2}{*}{AR} & \multirow{2}{*}{APs} & \multirow{2}{*}{ARs} \\
    \cline{1-2}
    \multicolumn{1}{c|}{PAF} & \multicolumn{1}{c|}{CM} & & & & \\
    \hline
    %
      1s, 10b, \hphantom{000-}256w & 1s, 10b, 256w & 65.8 & 70.3 & 56.1 & 61.1 \\
      2s, \hphantom{0}8b, 128-288w & 1s, \hphantom{0}8b, 256w & 66.1 & 70.5 & 56.7 & 61.9 \\
      2s, 10b, 128-256w & 1s, 10b, 256w & 66.1 & 70.7 & \textbf{57.0} & \textbf{62.0} \\
      3s, \hphantom{0}8b, \hphantom{0}96-256w & 1s, \hphantom{0}8b, 192w & \textbf{66.4} & \textbf{70.9} & 56.9 & 61.9 \\
      4s, \hphantom{0}8b, \hphantom{0}96-256w & 1s, \hphantom{0}8b, 224w & 65.7 & 70.2 & 56.3 & 61.4 \\
      5s, \hphantom{0}8b, \hphantom{0}64-256w & 1s, \hphantom{0}5b, 256w & 65.5 & 70.1 & 56.7 & 61.8 \\
    \hline
    \end{tabular}}
    \vspace{-5pt}
    \caption{Self-comparison on the body COCO validation set. All models have been tuned to have the same inference run-time. `APs' and `ARs' refer to the single-scale results.
    `PAF' represents the Part Affinity Field network configuration and `CM' the confidence map configuration. `s' refers to the number of stages of refinement, `b' to the number of convolutional blocks per stage, `w' to the number of output channels (or width) of each convolutional layer. All other settings follow Sec.~\ref{sec:param_config}.}
    \label{table:self_coco_val}
\end{table}

\subsection{Body and Foot Keypoint Detection Accuracy}\label{sec:body_val}
Once the optimal model 
has been selected, it is trained for whole-body estimation. Table~\ref{table:coco_val} show the accuracy results on the COCO validation set for our 3 different models, as well as the results reported by Cao~\etal~\cite{cao2018openpose}. 
The new deeper architecture slightly increases the accuracy of OpenPose when trained for whole-body estimation. It can also be applied to body-foot estimation, achieving a 1.1\% improvement in accuracy compared to that of Cao~\etal~\cite{cao2018openpose}.
Interestingly, adding face and hand keypoints to the same model results in a considerable decrease of the body detection accuracy by about 5\% for the shallow model when compared to that of \cite{cao2018openpose}. Intuitively, this is due to the fact that we are trying to fit nearly six times as many keypoints into the same network. 
The original model might not have enough capacity to handle the additional complexity introduced by the new keypoints.
However, this gap is smaller than 1\% for the improved architecture (deep body-foot vs. deep whole-body). The additional depth helps the network generalize to a higher number of output keypoints.

\begin{table}[h]
    \begin{center}
    \resizebox{\columnwidth}{!}{
    \begin{tabular}{l|c|c}
    \hline
    Method & Body AP  &	Foot AP \\
    \hline
    { Body-foot OpenPose (multi-scale)~\cite{cao2018openpose}} & 65.3 & \textbf{77.9} \\
    { Shallow whole-body (ours, multi-scale)} & 60.9 & 70.2 \\
    { Deep body-foot (ours, multi-scale)} & \textbf{66.4} & 76.8 \\
    { Deep whole-body (ours, multi-scale)} & 65.6 & 76.2 \\
    \hline
    \end{tabular}}
    \end{center}
    \vspace{-5pt}
    \caption{Accuracy results on the COCO
    validation set. 
    `Shallow' refers to the network architecture with the same depth and input resolution as that of OpenPose, while `Deep' refers to our improved architecture. `Body-foot' refers to the network that simply predicts body and foot keypoints, following the default OpenPose output, while `Whole-body' refers to our novel single-network model.} 
    \label{table:coco_val}
\end{table}


\subsection{Face Keypoint Detection Accuracy}\label{sec:face_val}
In order to evaluate the accuracy of face alignment, traditional approaches have used the Probability of Correct Keypoint (PCK) metric,
which checks the probability that a predicted keypoint is within a distance threshold of its true location.
However, it 
does not generalize to a multi-person setting.
In order to evaluate our work, we reuse the mean Average Precision (AP) and Recall (AR), following the COCO multi-person metric. 
We train our whole-body algorithm with the same face datasets that OpenPose~\cite{cao2018openpose} used: Multi-PIE~\cite{gross2010multi}, FRGC~\cite{frgc_2010}, and i-bug~\cite{sagonas2016300}.
We create a custom validation set by selecting a small subset of images from each dataset.
We show the results in Table~\ref{table:face_val}. 
We can see that both our method and \cite{cao2018openpose} greatly over-fit on the Multi-PIE and FRGC datasets. These datasets consist of images annotated in controlled lab environments, and all faces appear frontal and with no occlusion, similar to the last image in Fig.~\ref{fig:dataset_examples}. However, their accuracy is considerably lower in the in-the-wild i-bug dataset, where our approach is about 2\% more accurate. 

\begin{table}[h]
    \begin{center}
    \setlength{\tabcolsep}{3pt}
    \resizebox{\columnwidth}{!}{
    \begin{tabular}{l|c|c|c}
    \hline
    \multirow{2}{*}{ Method } & \multicolumn{3}{c}{Face AR}\\
    \cline{2-4}
    & FRGC & M-Pie & i-bug \\
    \hline
    { OpenPose~\cite{cao2018openpose}} & 98.3 & \textbf{96.3} & 52.4 \\
    { Shallow Whole-body (ours, single-scale)} & \textbf{98.4} & 90.6 & 50.6 \\ 
    { Deep Whole-body (ours, single-scale)} & \textbf{98.4} & 93.2 & \textbf{54.5} \\
    \hline
    \end{tabular}}
    \end{center}
    \vspace{-5pt}
    \caption{Accuracy results on our custom CMU Multi-PIE and FRGC validation sets. 
    All the people in each image are not necessarily labeled on i-bug. Thus, those samples might be considered erroneous ``false positives" and affect the AP results. However, AR is only affected by the annotated samples, so it is used as the main metric for i-bug.}
    \label{table:face_val}
\end{table}

\subsection{Hand Keypoint Detection Accuracy}\label{sec:hand_val}
Analog to face evaluation, we randomly select a subset of images from each hand dataset for validation. We denote ``Hand Dome" for the subset of~\cite{simon2017hand} recorded in the Panoptic Studio~\cite{Joo_2017_TPAMI}, and ``Hand MPII" for the subset manually annotated from MPII~\cite{andriluka14cvpr} images.
The results are presented in Table~\ref{table:hand_val}. Both our method and the previous OpenPose over-fit on the Dome dataset, where usually only a single person appears in each frame, similar to the first image in Fig.~\ref{fig:dataset_examples}. However, the manually annotated images from MPII are more challenging for both approaches, as it represents realistic in-the-wild scenes. In such images, we can see the clear benefit of our deeper network with respect to Cao~\etal~\cite{cao2018openpose} and our initial shallow model, outperforming by about 5.5\% on the Hand MPII dataset. 

\begin{table}[b]
    \begin{center}
    \resizebox{\columnwidth}{!}{
    \begin{tabular}{l|c|c}
    \hline
    \multirow{2}{*}{ Method } & \multicolumn{2}{c}{Hand AR}\\
    \cline{2-3}
    & { Dome}  &	{ MPII } \\
    \hline
    { OpenPose (single-scale)~\cite{cao2018openpose}} & 97.0 & 82.7 \\
    { Shallow whole-body (ours, single-scale)} & 94.6 & 82.4 \\
    { Deep whole-body (ours, single-scale)} & \textbf{97.8} & \textbf{88.1} \\
    \hline
    \end{tabular}}
    \end{center}
    \vspace{-5pt}
    \caption{Results on our custom Hand~Dome and Hand~MPII validation sets. These datasets might contain unlabeled people (similar to i-bug), so AR is used for evaluation.}
    \label{table:hand_val}
\end{table}


\begin{figure}[t]
\centering
    \includegraphics[width=0.85\linewidth]{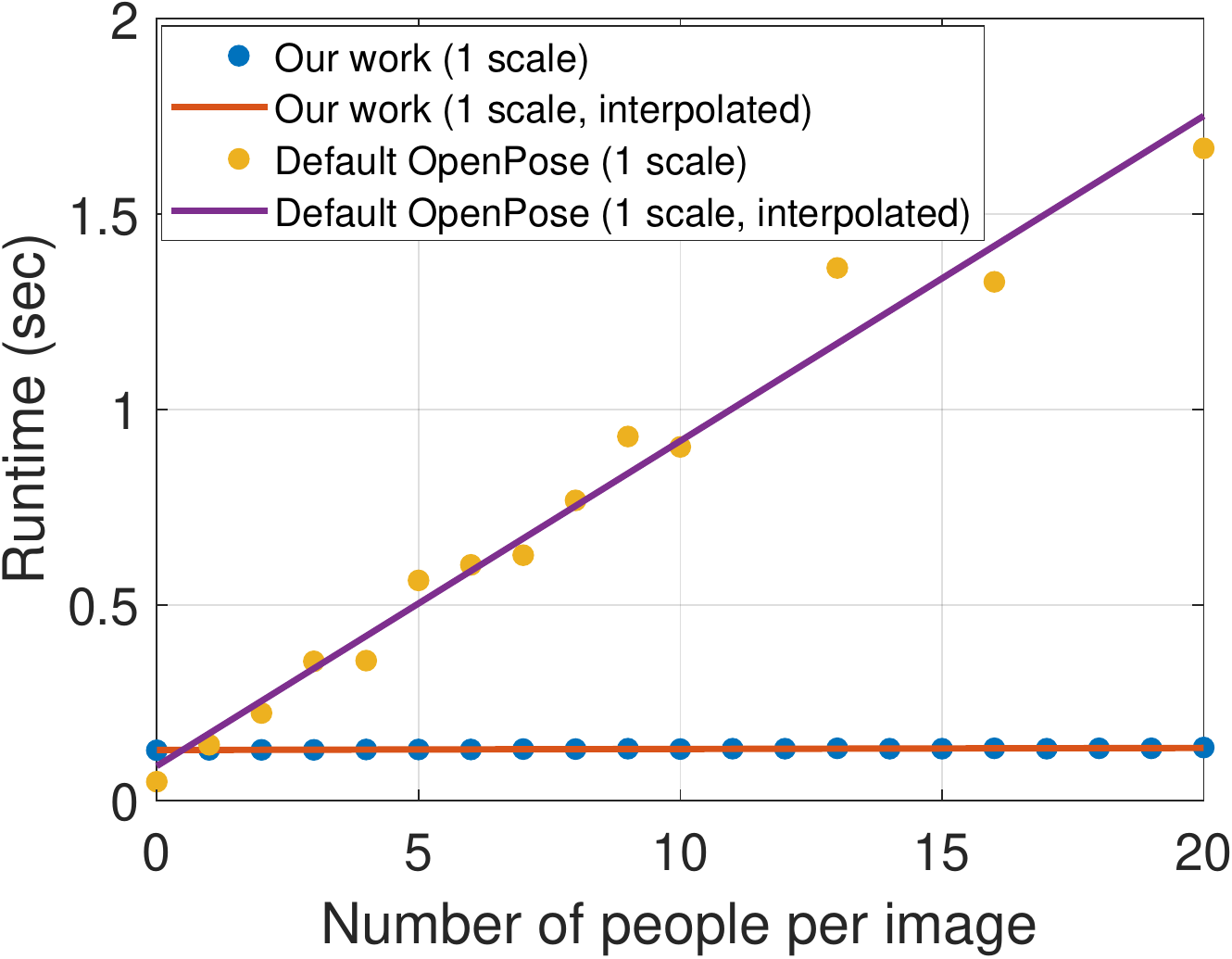}
    \vspace{-5pt}
    \caption{Inference time comparison between our work and whole-body OpenPose~\cite{cao2018openpose}. While the inference time of our proposed approach is invariant, the run-time of OpenPose grows linearly with the number of people.
    OpenPose run-time presents some oscillations because it does not run face and hand detectors if the nose or wrist keypoints (provided by the body network) of a person are not found. This is a common case in images with many crowded images.
    This analysis was performed on a system with a Nvidia 1080 Ti. 
    }
\label{fig:runtime_comparison}
\end{figure}

\subsection{Run-time Comparison}\label{sec:runtime_comparison}
In Fig.~\ref{fig:runtime_comparison}, we compare the inference run-time between OpenPose and our work.
Our method is 10\% faster than OpenPose for images with a single person. However, the inference time of our single-network approach remains constant, while OpenPose's time is proportional to the number of people detected, in particular, it is proportional to the number of face and hand proposals. 
This leads to a massive speedup of our approach
when the number of people increases. For images with $n$ people, our approach is approximately $n$ times faster than OpenPose. For crowded images, many hands and faces are occluded, slightly reducing this speedup.
For instance, our approach is about 7 times faster than OpenPose for typical images with 10 people in them.
\begin{figure*}[ht]
\centering
    \begin{subfigure}[t]{1.\textwidth}
        \begin{subfigure}[t]{0.195\textwidth}
        \includegraphics[width=1\linewidth]{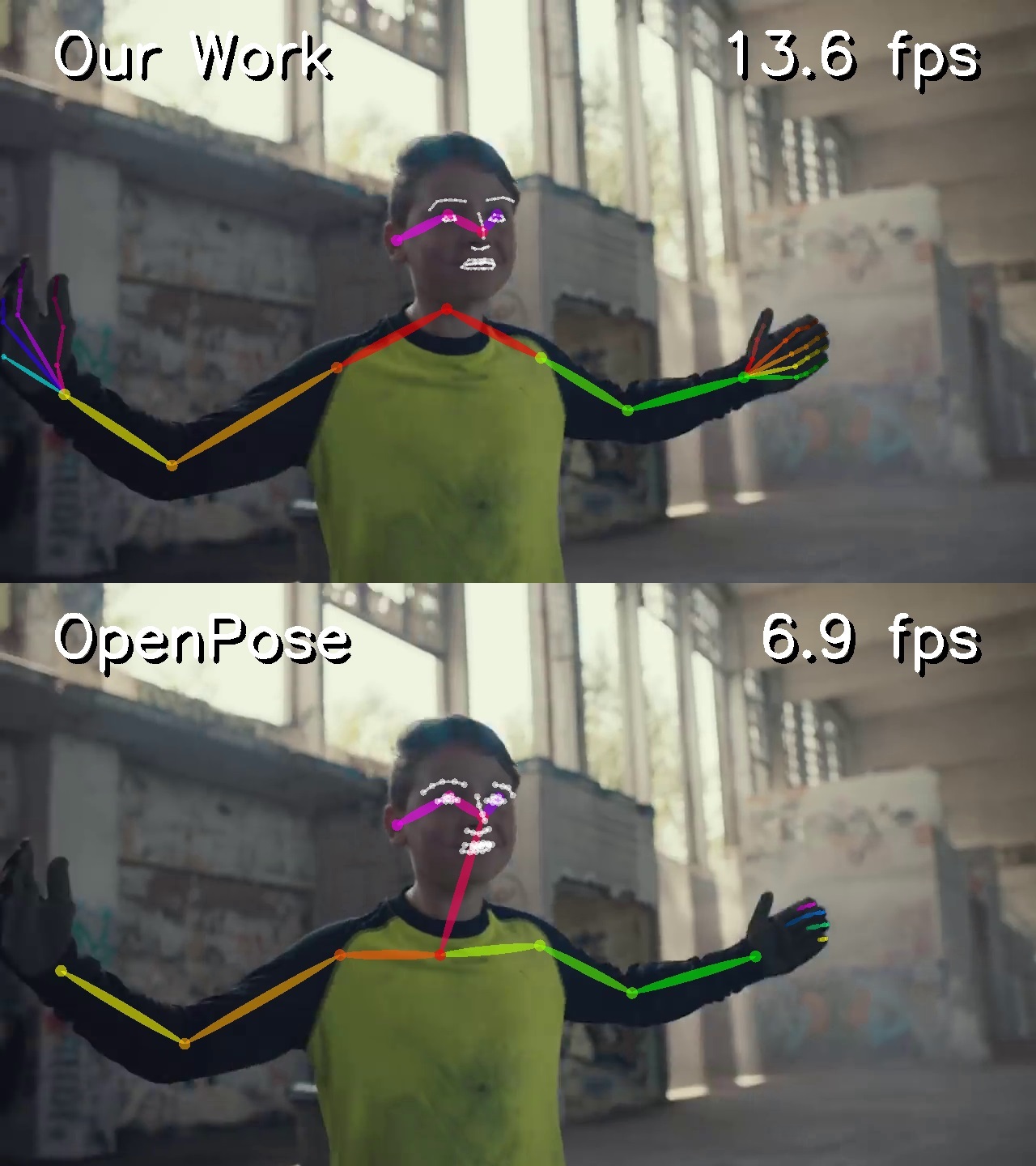}
        \vspace{-17.5pt}
        \caption{}
        \end{subfigure}
        \begin{subfigure}[t]{0.195\textwidth}
        \includegraphics[width=1\linewidth]{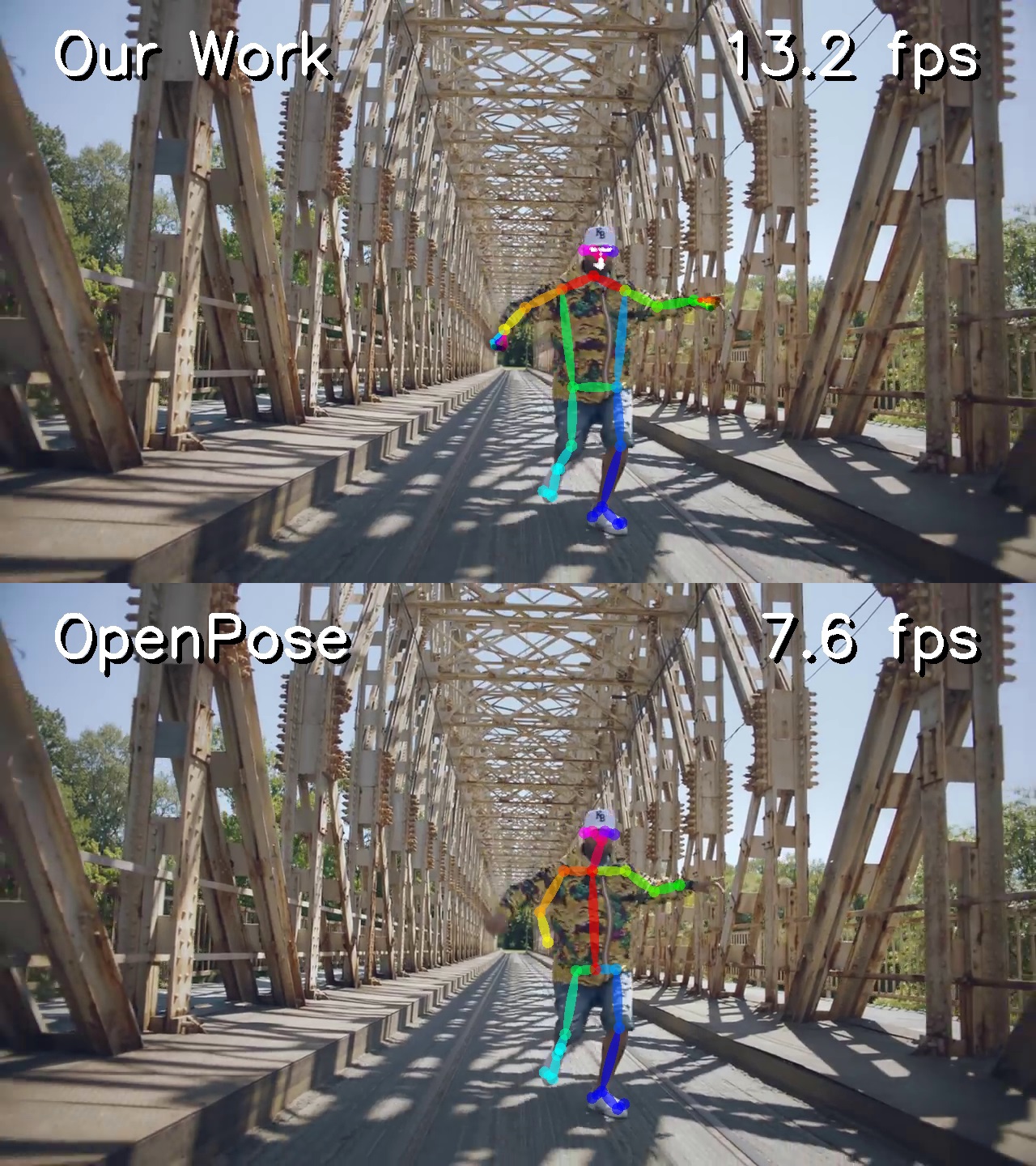}
        \vspace{-17.5pt}
        \caption{}
        \end{subfigure}
        \begin{subfigure}[t]{0.195\textwidth}
        \includegraphics[width=1\linewidth]{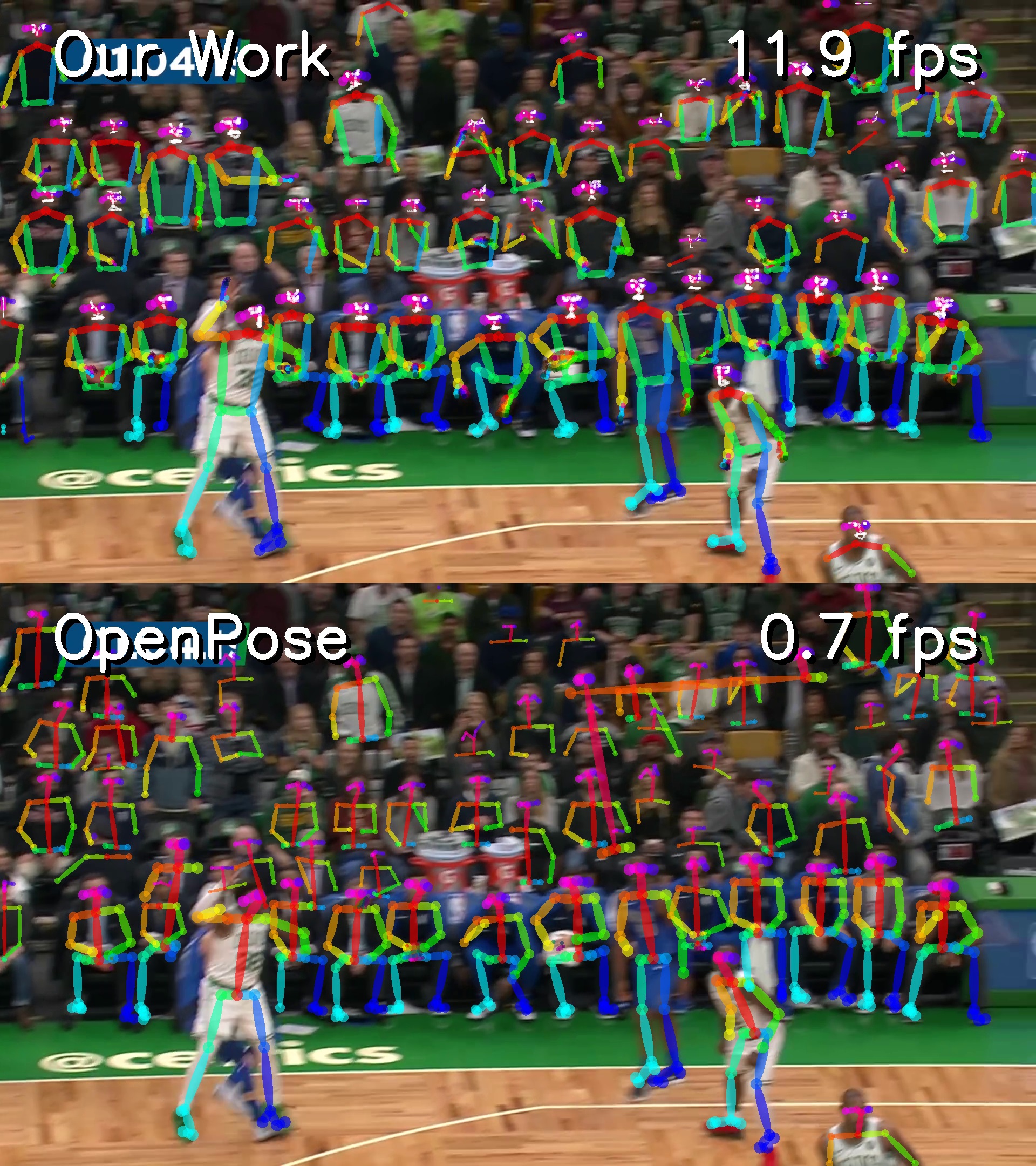}
        \vspace{-17.5pt}
        \caption{}
        \end{subfigure}
        \begin{subfigure}[t]{0.195\textwidth}
        \includegraphics[width=1\linewidth]{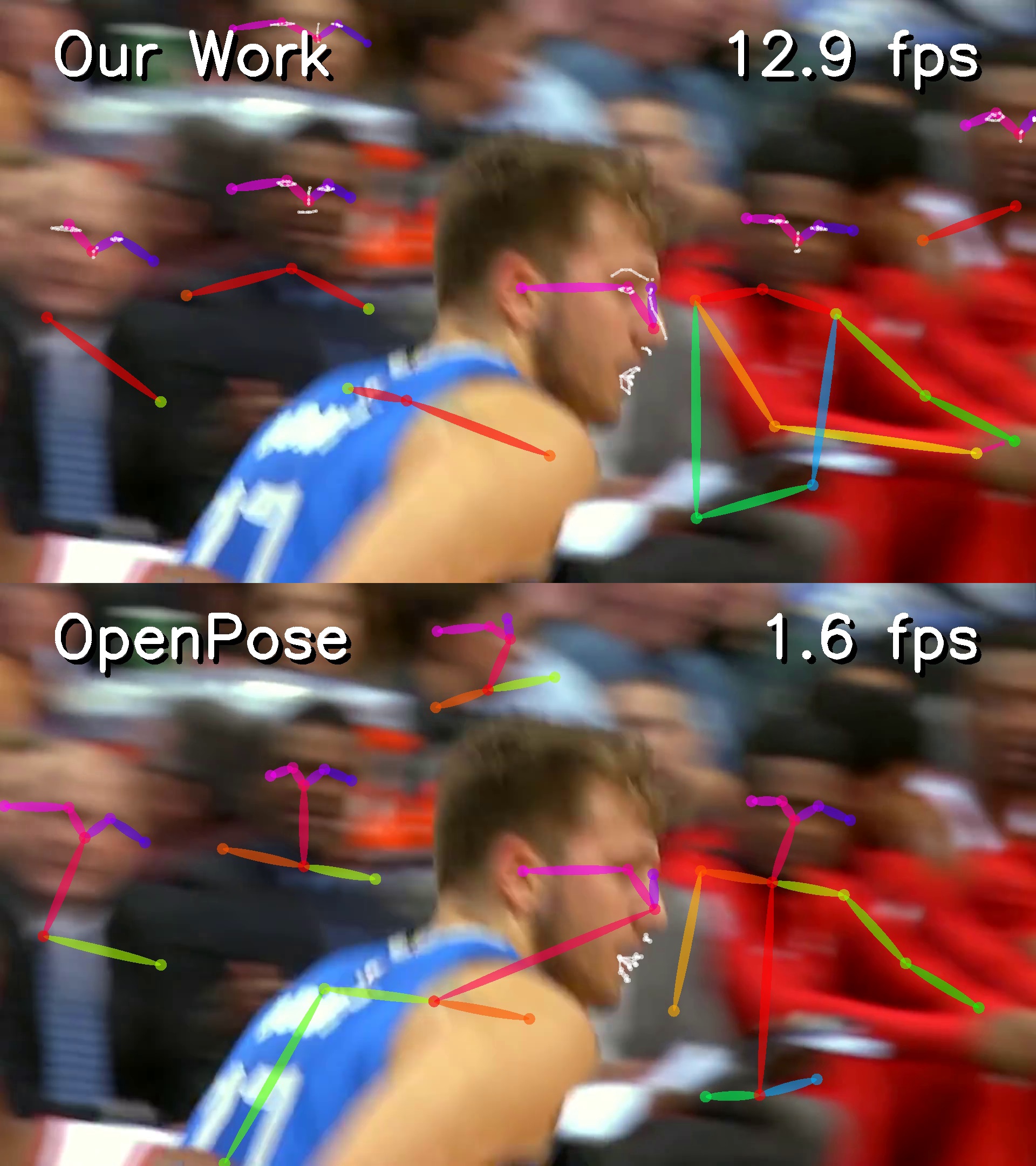}
        \vspace{-17.5pt}
        \caption{}
        \end{subfigure}
        \begin{subfigure}[t]{0.195\textwidth}
        \includegraphics[width=1\linewidth]{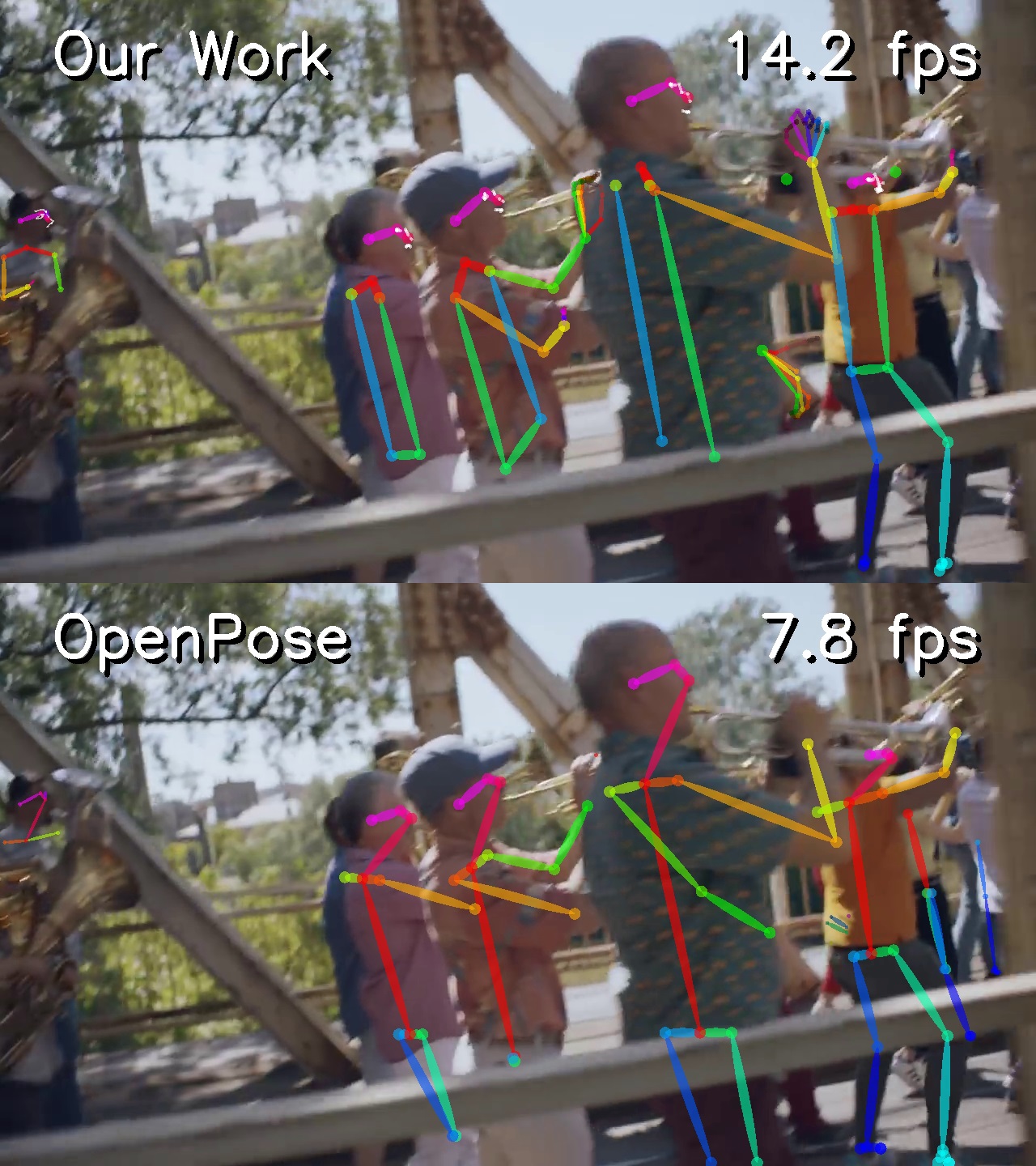}
        \vspace{-17.5pt}
        \caption{}
        \end{subfigure}
    \end{subfigure}
    \begin{subfigure}[t]{1.\textwidth}
        \begin{subfigure}[t]{0.195\textwidth}
        \includegraphics[width=1\linewidth]{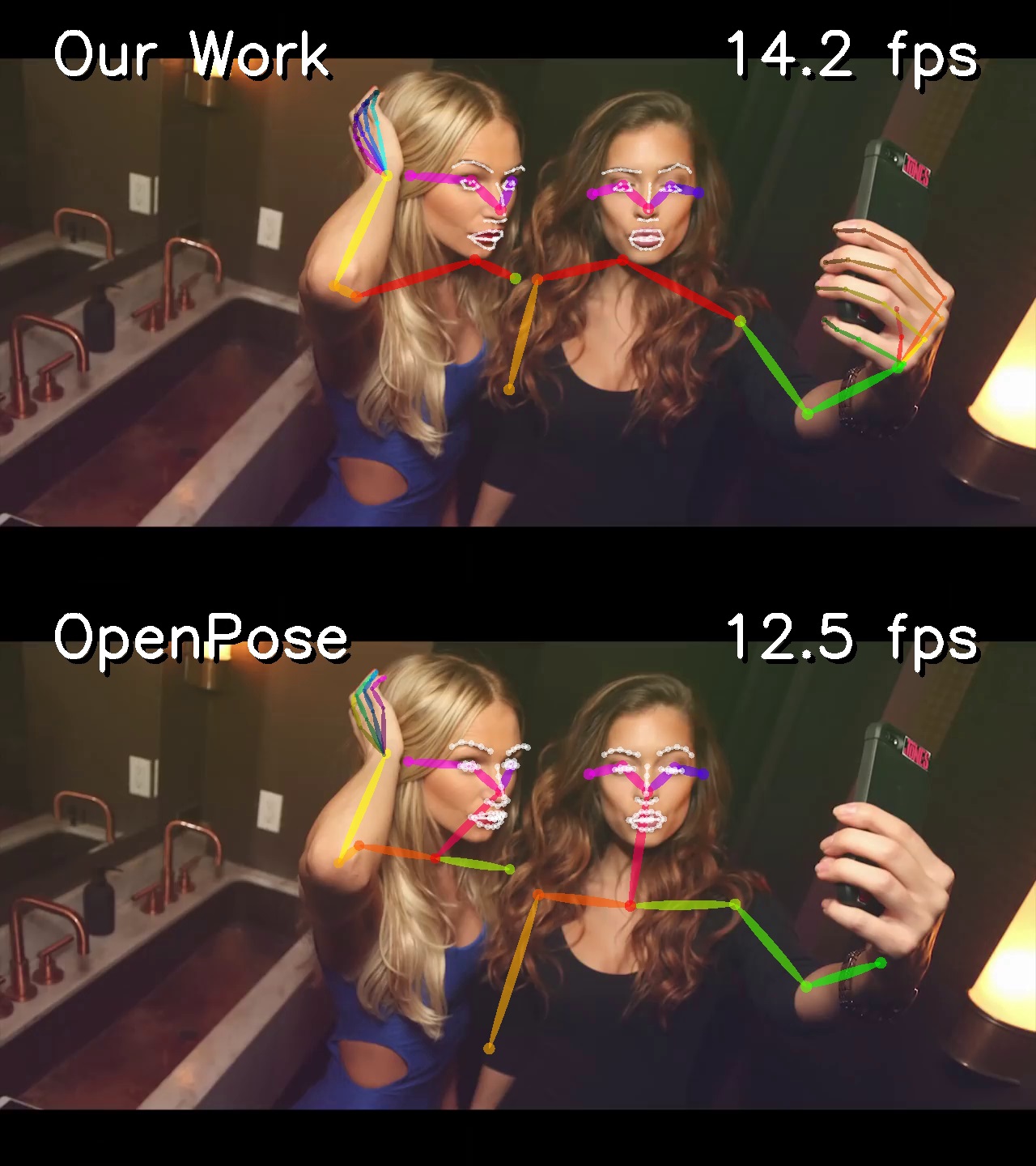}
        \vspace{-17.5pt}
        \caption{}
        \end{subfigure}
        \begin{subfigure}[t]{0.195\textwidth}
        \includegraphics[width=1\linewidth]{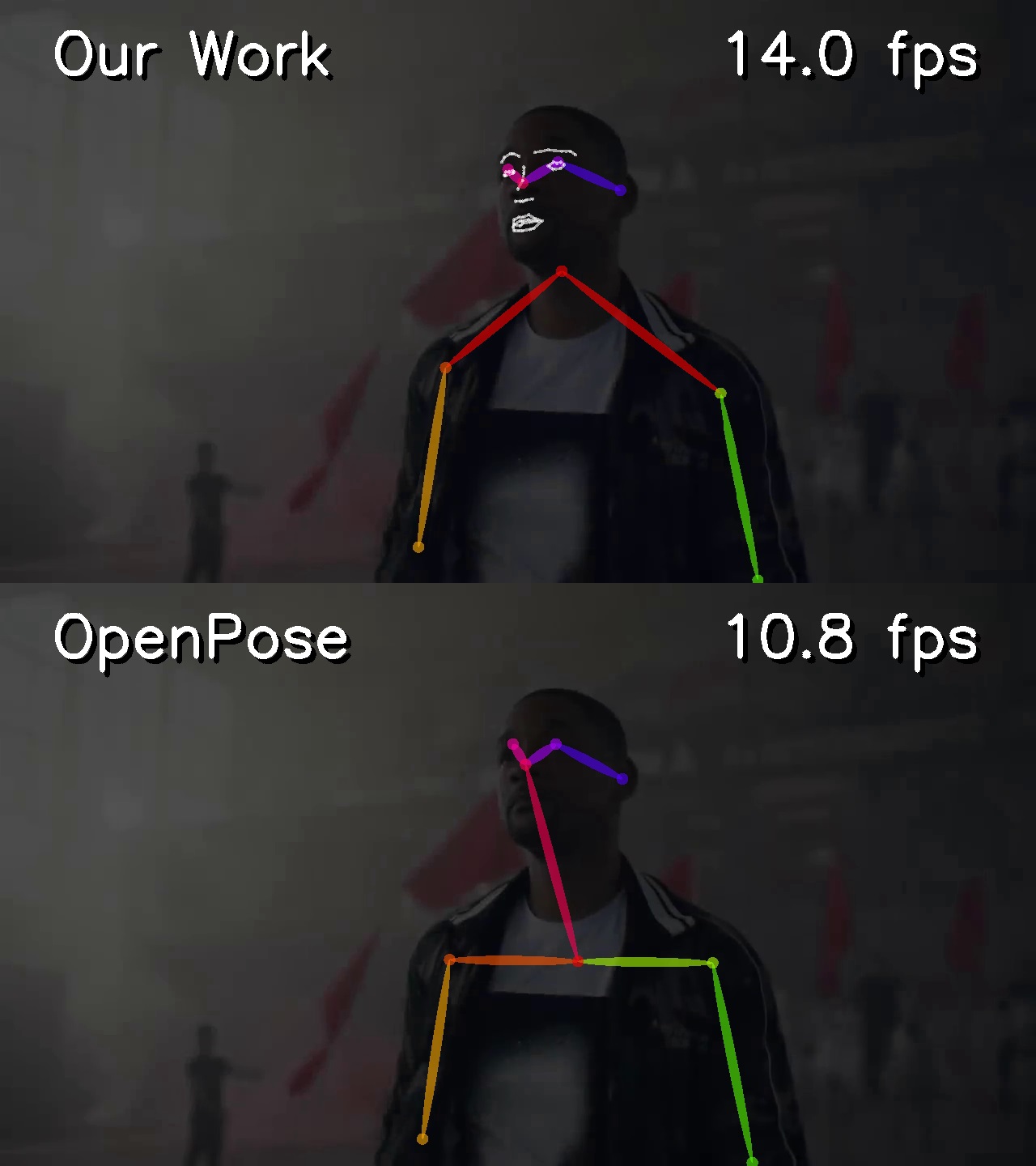}
        \vspace{-17.5pt}
        \caption{}
        \end{subfigure}
        \begin{subfigure}[t]{0.195\textwidth}
        \includegraphics[width=1\linewidth]{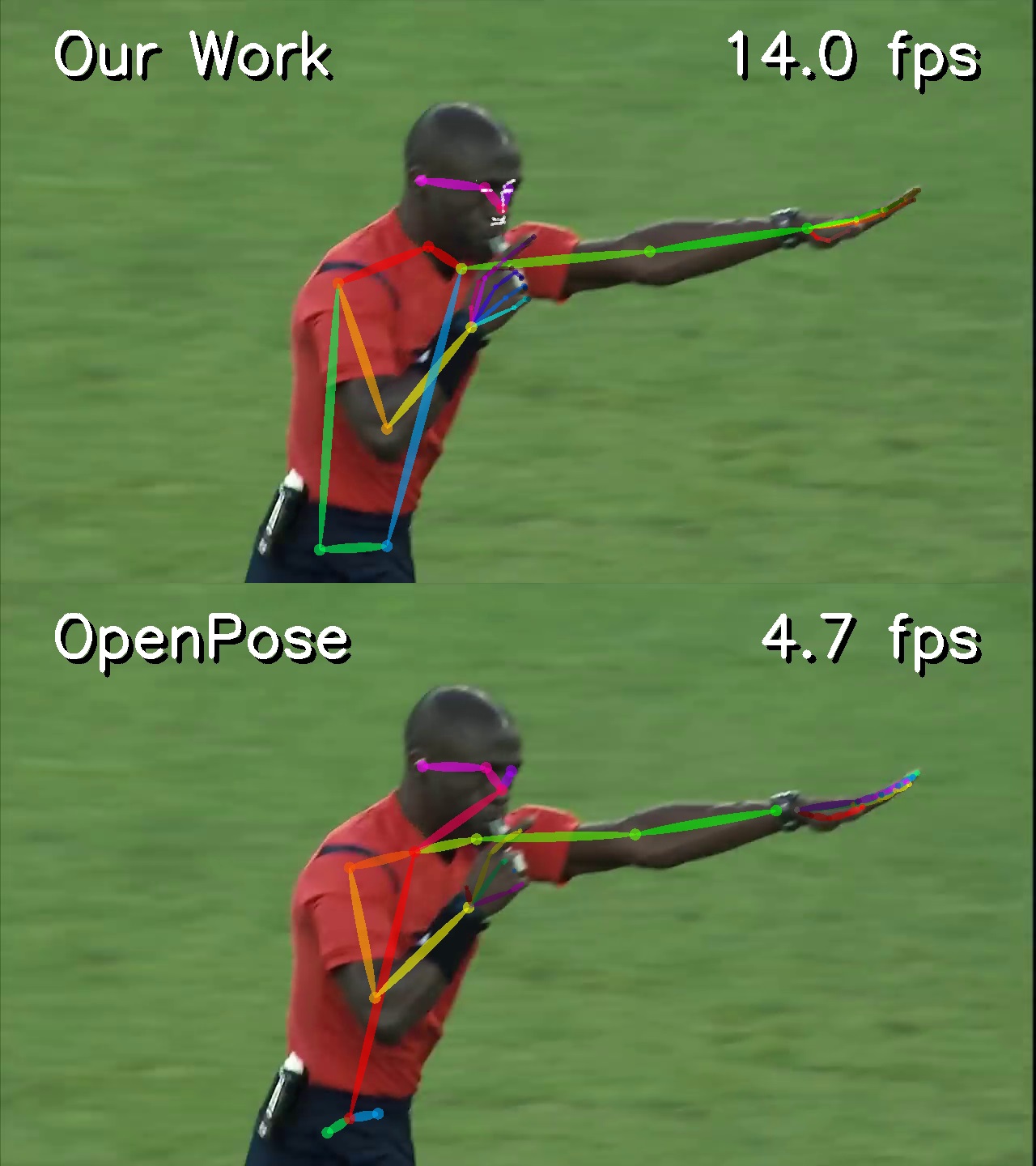}
        \vspace{-17.5pt}
        \caption{}
        \end{subfigure}
        \begin{subfigure}[t]{0.195\textwidth}
        \includegraphics[width=1\linewidth]{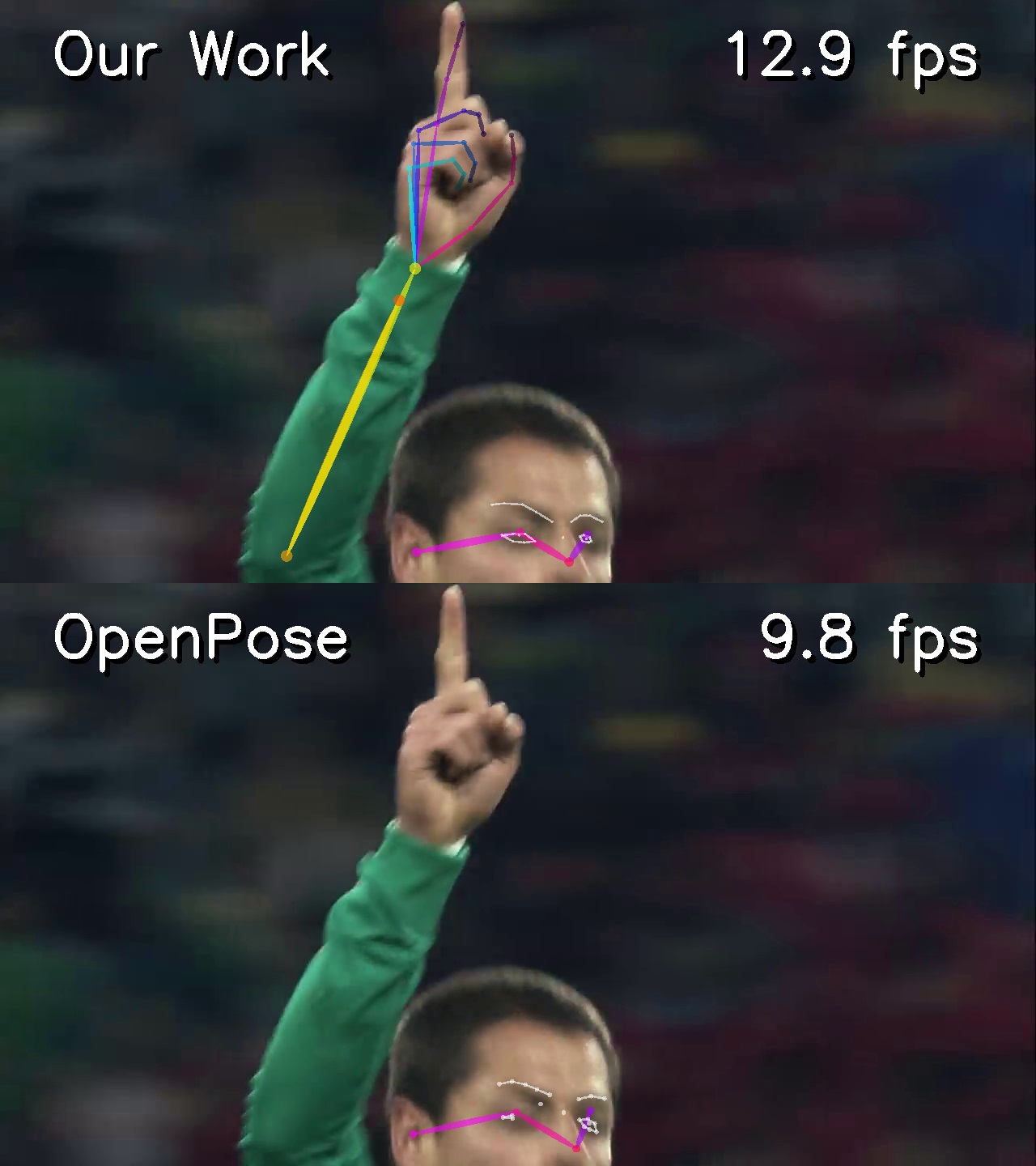}
        \vspace{-17.5pt}
        \caption{}
        \end{subfigure}
        \begin{subfigure}[t]{0.195\textwidth}
        \includegraphics[width=1\linewidth]{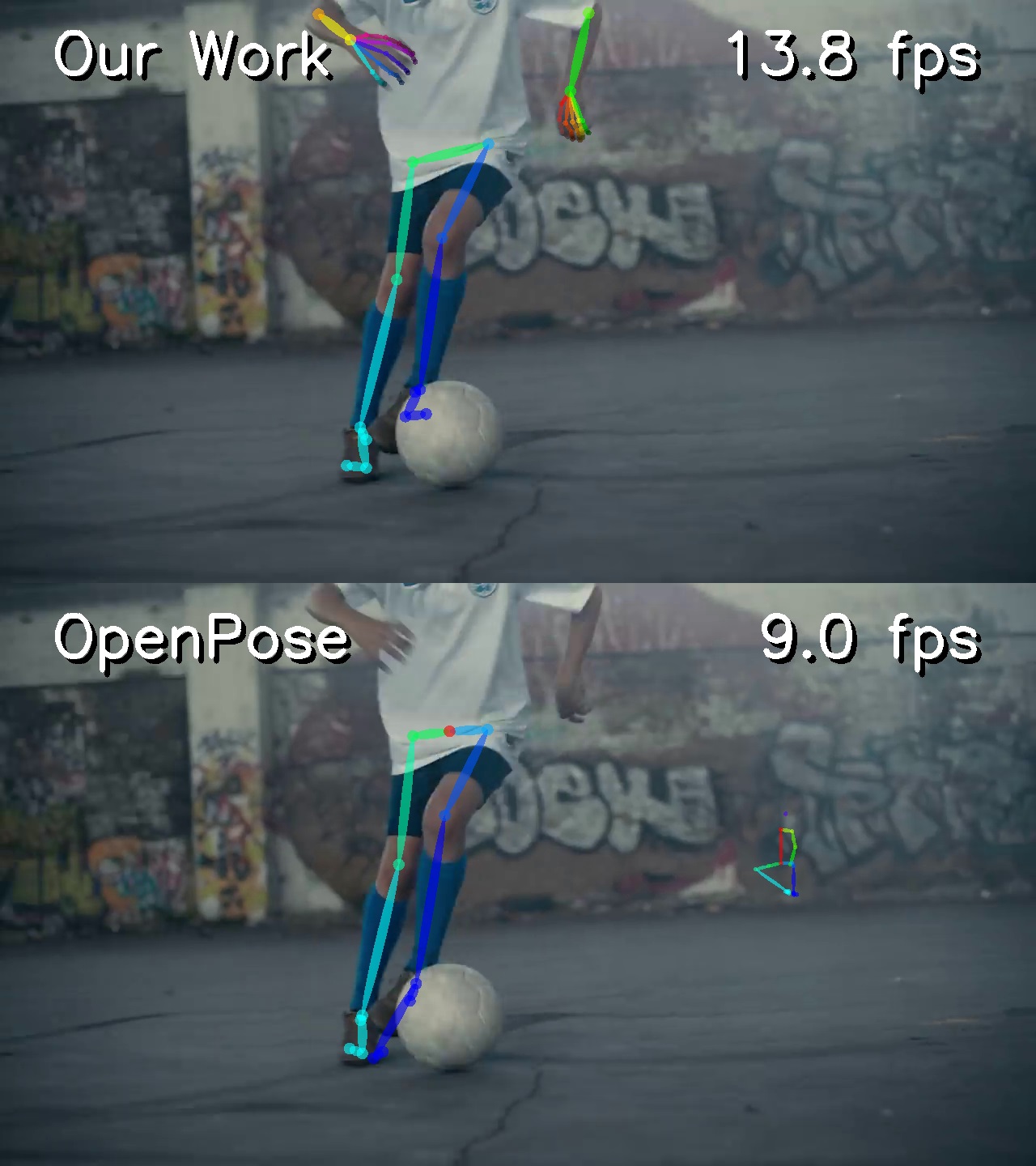}
        \vspace{-17.5pt}
        \caption{}
        \end{subfigure}
    \end{subfigure}
    \begin{subfigure}[t]{1.\textwidth}
        \begin{subfigure}[t]{0.195\textwidth}
        \includegraphics[width=1\linewidth]{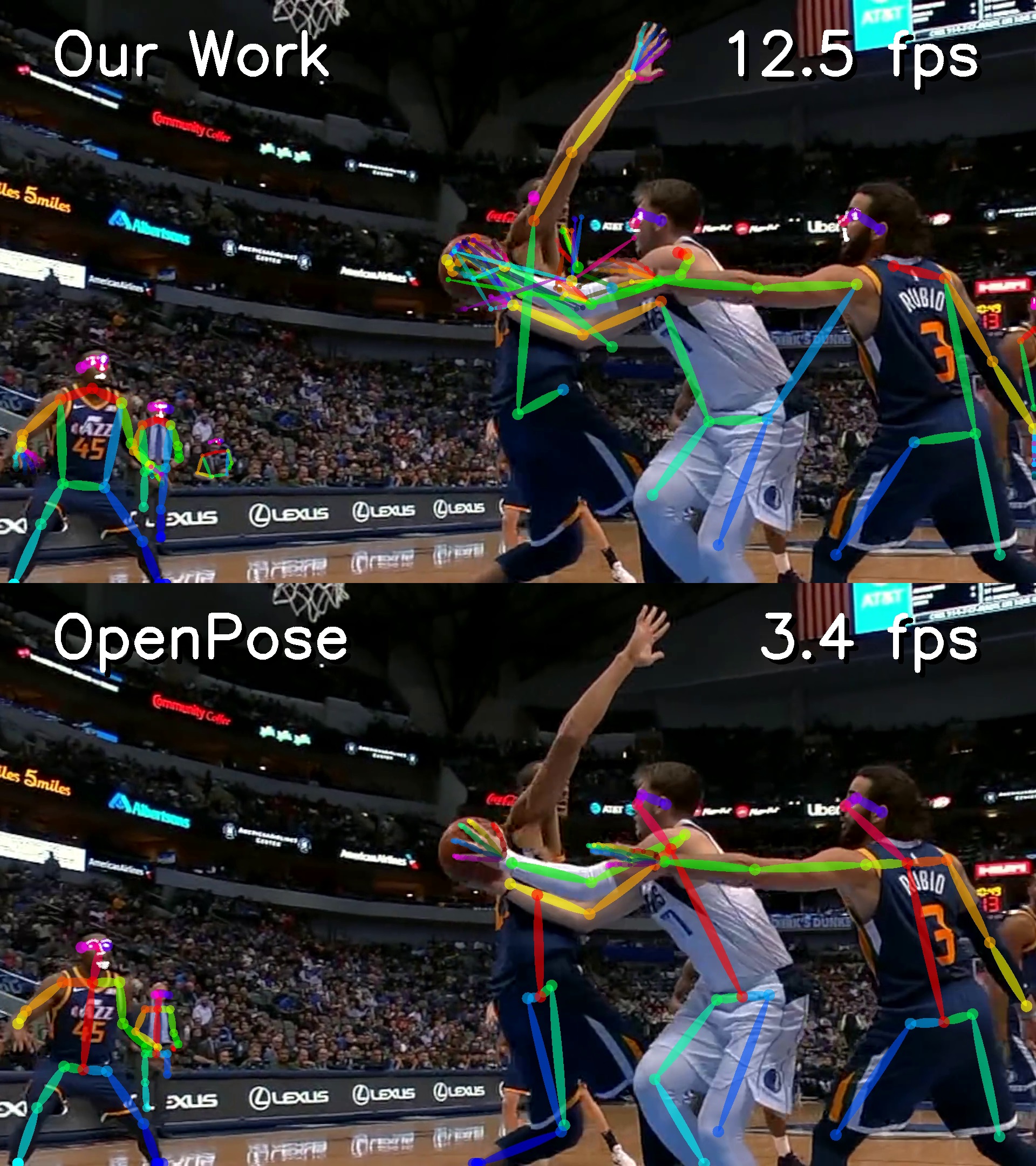}
        \vspace{-17.5pt}
        \caption{}
        \end{subfigure}
        \begin{subfigure}[t]{0.195\textwidth}
        \includegraphics[width=1\linewidth]{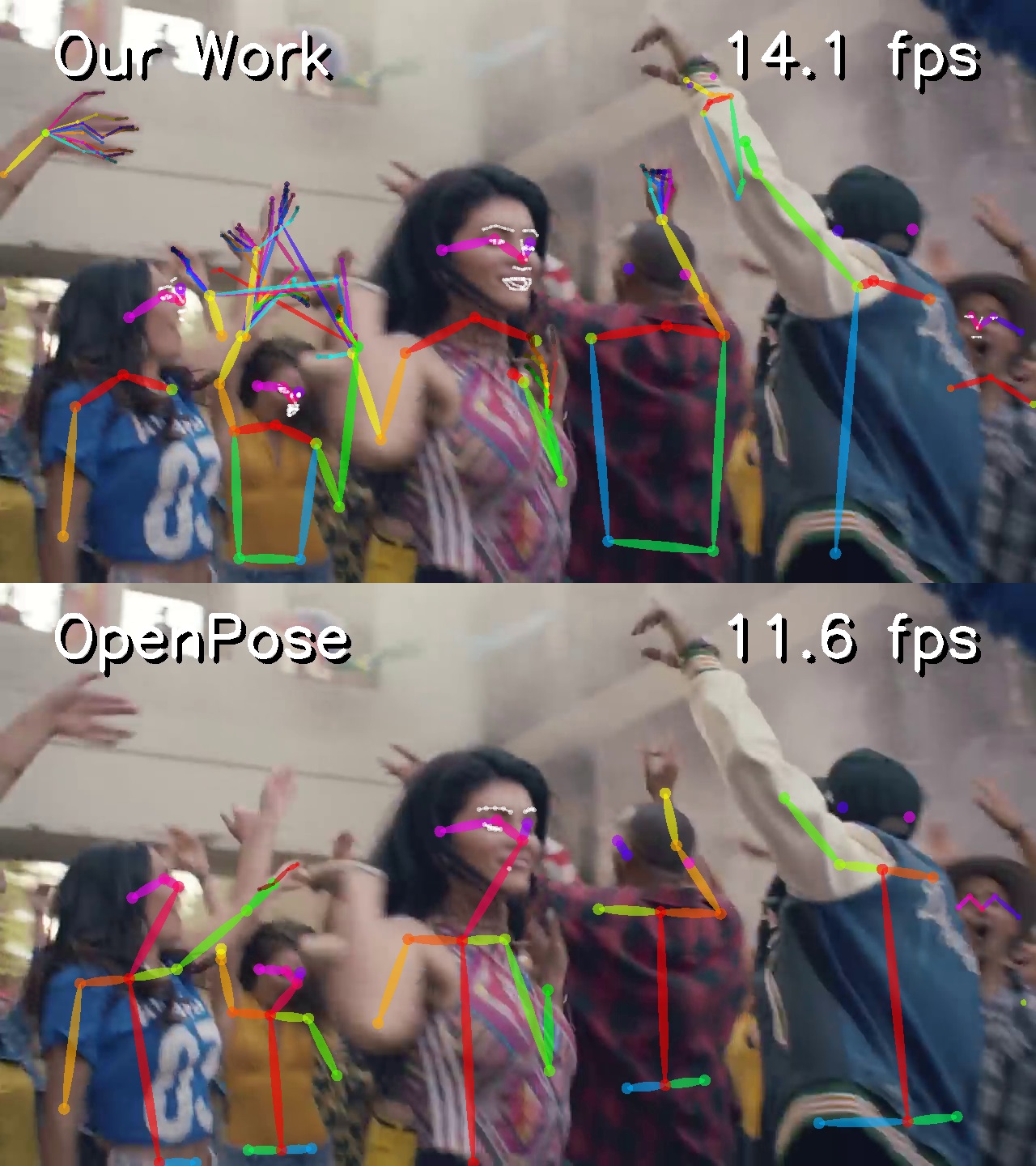}
        \vspace{-17.5pt}
        \caption{}
        \end{subfigure}
        \begin{subfigure}[t]{0.195\textwidth}
        \includegraphics[width=1\linewidth]{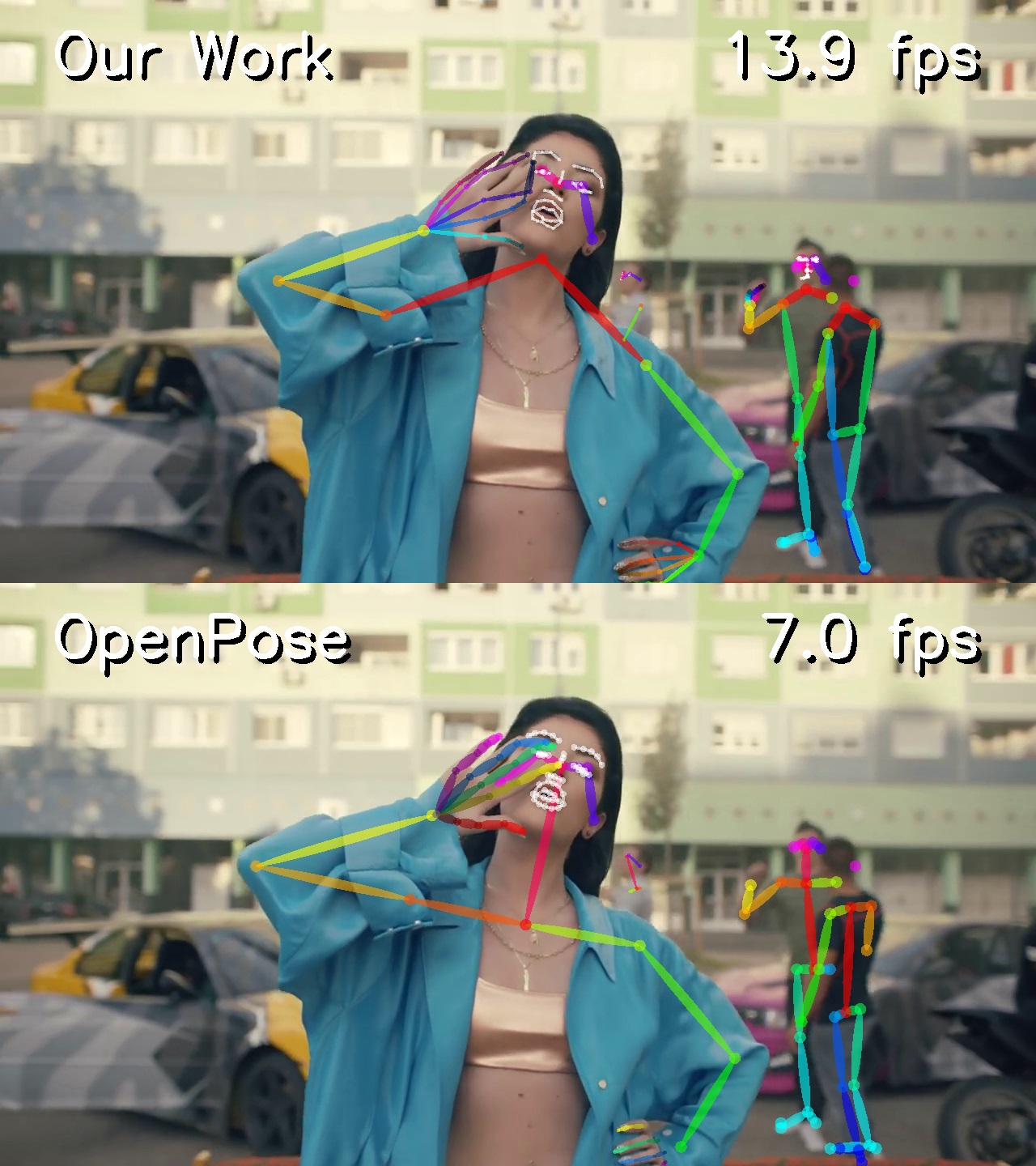}
        \vspace{-17.5pt}
        \caption{}
        \end{subfigure}
        \begin{subfigure}[t]{0.195\textwidth}
        \includegraphics[width=1\linewidth]{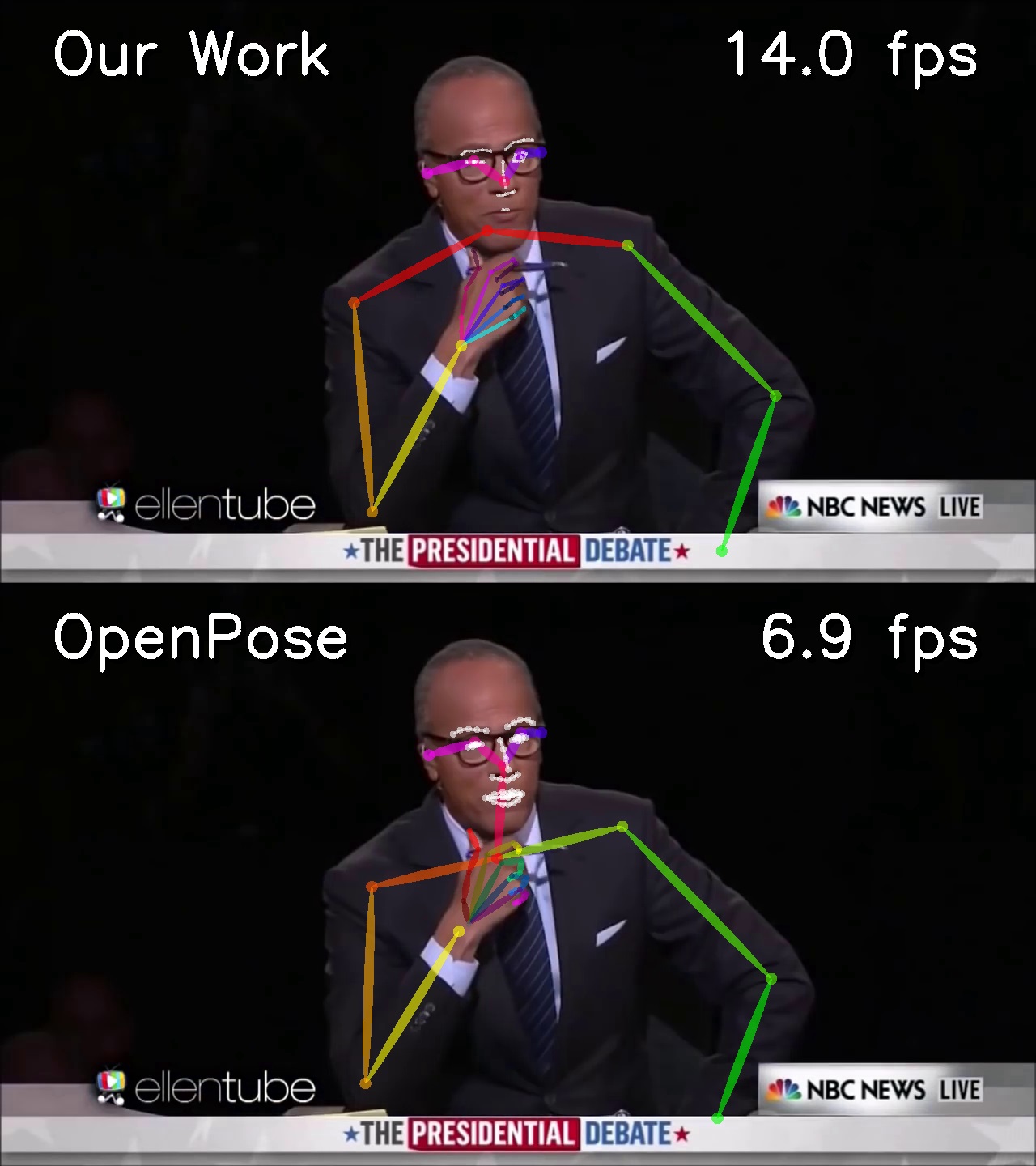}
        \vspace{-17.5pt}
        \caption{}
        \end{subfigure}
        \begin{subfigure}[t]{0.195\textwidth}
        \includegraphics[width=1\linewidth]{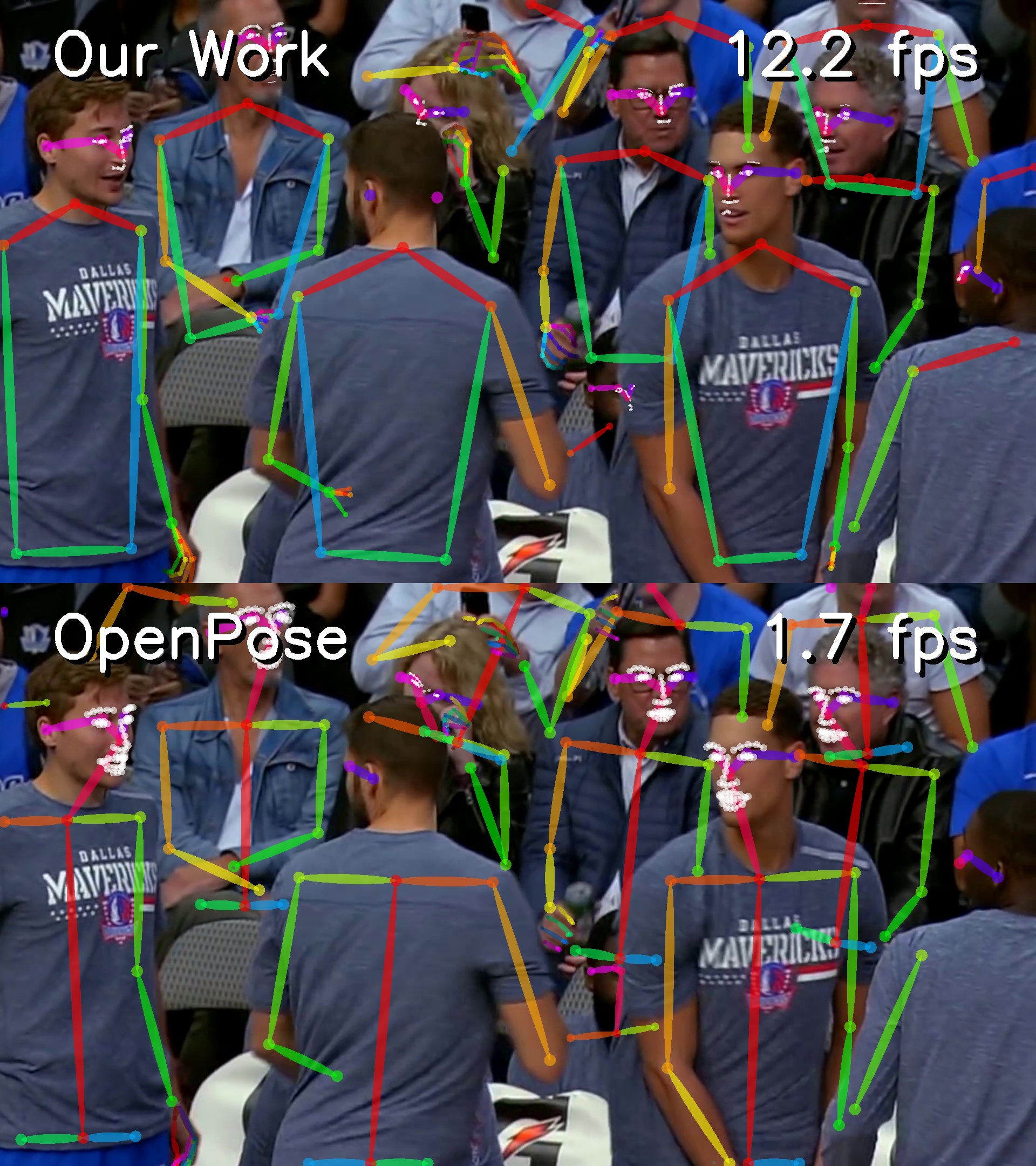}
        \vspace{-17.5pt}
        \caption{}
        \end{subfigure}
    \end{subfigure}
    \vspace{-10pt}
    \caption{Qualitative comparison between our method (top) and the previous version of OpenPose~\cite{cao2018openpose} (bottom), performed on a system with 2 Nvidia 1080 Ti.
    (a-j)~show improved results and (k-o) failing cases.
    (a)~It generalizes better to hands wearing any kind of gloves.
    (b)~The implicit finger information helps wrist and elbow detection.
    (c)~Even smaller faces and hands are detected.
    (d-e)~Blurry and profile faces are detected.
    (f)~More extreme hand poses are detected.
    (g)~Faces from low-brightness images are better detected.
    (h)~Hands where all fingers are occluded are detected.
    (i-j)~Cropped arms are properly detected.
    (k-l)~It shows difficulties when several hands are in proximity.
    (m-o)~It seems to fail for some relatively easy hand and face poses that are successfully detected with the previous version of OpenPose.
    }\label{fig:qualitative_results}
\end{figure*}

\section{Conclusion}
In this paper, we resort to multi-task learning combined with an improved model architecture to train the first single-network approach for 2D whole-body estimation. 
Our work brings together multiple and, currently, independent keypoint detection tasks into a unified framework.
We evaluate our method on multiple keypoint detection benchmarks and compare it with the state-of-the-art, considerably outperforming it in both training and testing speed as well as slightly improving its accuracy. 
We qualitatively show in Fig.~\ref{fig:qualitative_results} that our face and hand detectors generalize better to in-the-wild images, 
benefiting from their indirect exposure to the immense body datasets. 
Nevertheless, there are still some limitations with our method. First, we observe global failure cases when a significant part of the target person is occluded or outside of the image boundaries. Secondly, the accuracy of the face and especially hand keypoint detectors is still limited, failing in the case of severe motion blur, small people, and extreme gestures. 
Third, we qualitatively observe that the previous version of OpenPose outperforms ours for face and hand detection when poses are simple and no occlusion occurs. The previous method crops the bounding box proposal of those bounding box candidates, resizes them up, and feeds them into its dedicated networks. This higher input resolution leads to an increased pixel localization precision if the keypoint detection is successful.


{\small
\bibliographystyle{ieee_fullname} 
\interlinepenalty=10000
\bibliography{iccv_cam_ready} 

\begin{thebibliography}{10}\itemsep=-1pt

\bibitem{alp2018densepose}
R{\i}za Alp~G{\"u}ler, Natalia Neverova, and Iasonas Kokkinos.
\newblock Densepose: Dense human pose estimation in the wild.
\newblock In {\em CVPR}, pages 7297--7306, 2018.

\bibitem{andriluka20142d}
Mykhaylo Andriluka, Leonid Pishchulin, Peter Gehler, and Bernt Schiele.
\newblock {2D} human pose estimation: new benchmark and state of the art
  analysis.
\newblock In {\em CVPR}, 2014.

\bibitem{andriluka14cvpr}
Mykhaylo Andriluka, Leonid Pishchulin, Peter Gehler, and Bernt Schiele.
\newblock 2d human pose estimation: New benchmark and state of the art
  analysis.
\newblock In {\em CVPR}, June 2014.

\bibitem{Andriluka2010}
Mykhaylo Andriluka, Stefan Roth, and Bernt Schiele.
\newblock {M}onocular 3{D} pose estimation and tracking by detection.
\newblock In {\em CVPR}, 2010.

\bibitem{asthana2014incremental}
Akshay Asthana, Stefanos Zafeiriou, Shiyang Cheng, and Maja Pantic.
\newblock Incremental face alignment in the wild.
\newblock In {\em CVPR}, pages 1859--1866, 2014.

\bibitem{Recycle-GAN}
Aayush Bansal, Shugao Ma, Deva Ramanan, and Yaser Sheikh.
\newblock Recycle-gan: Unsupervised video retargeting.
\newblock In {\em ECCV}, 2018.

\bibitem{belagiannis2017recurrent}
Vasileios Belagiannis and Andrew Zisserman.
\newblock Recurrent human pose estimation.
\newblock In {\em IEEE FG}, 2017.

\bibitem{cai2018weakly}
Yujun Cai, Liuhao Ge, Jianfei Cai, and Junsong Yuan.
\newblock Weakly-supervised 3d hand pose estimation from monocular rgb images.
\newblock In {\em ECCV}, pages 666--682, 2018.

\bibitem{cao2018openpose}
Zhe Cao, Gines Hidalgo, Tomas Simon, Shih-En Wei, and Yaser Sheikh.
\newblock Open{P}ose: realtime multi-person 2{D} pose estimation using {P}art
  {A}ffinity {F}ields.
\newblock In {\em arXiv preprint arXiv:1812.08008}, 2018.

\bibitem{cao2017realtime}
Zhe Cao, Tomas Simon, Shih-En Wei, and Yaser Sheikh.
\newblock Realtime multi-person 2d pose estimation using part affinity fields.
\newblock In {\em CVPR}, 2017.

\bibitem{chan2018everybody}
Caroline Chan, Shiry Ginosar, Tinghui Zhou, and Alexei~A Efros.
\newblock Everybody dance now.
\newblock In {\em ECCV Workshop}, 2018.

\bibitem{chen2017adversarial}
Yu Chen, Chunhua Shen, Xiu-Shen Wei, Lingqiao Liu, and Jian Yang.
\newblock Adversarial posenet: A structure-aware convolutional network for
  human pose estimation.
\newblock In {\em ICCV}, 2017.

\bibitem{chen2017cascaded}
Yilun Chen, Zhicheng Wang, Yuxiang Peng, Zhiqiang Zhang, Gang Yu, and Jian Sun.
\newblock Cascaded pyramid network for multi-person pose estimation.
\newblock In {\em CVPR}, 2018.

\bibitem{chu2017multi}
Xiao Chu, Wei Yang, Wanli Ouyang, Cheng Ma, Alan~L Yuille, and Xiaogang Wang.
\newblock Multi-context attention for human pose estimation.
\newblock In {\em CVPR}, 2017.

\bibitem{Dantone2013}
Matthias Dantone, Juergen Gall, Christian Leistner, and Luc Van~Gool.
\newblock Human pose estimation using body parts dependent joint regressors.
\newblock In {\em CVPR}, 2013.

\bibitem{duong2015low}
Long Duong, Trevor Cohn, Steven Bird, and Paul Cook.
\newblock Low resource dependency parsing: Cross-lingual parameter sharing in a
  neural network parser.
\newblock In {\em ACL-IJCNLP}, volume~2, pages 845--850, 2015.

\bibitem{fang2017rmpe}
Hao-Shu Fang, Shuqin Xie, Yu-Wing Tai, and Cewu Lu.
\newblock {RMPE}: Regional multi-person pose estimation.
\newblock In {\em ICCV}, 2017.

\bibitem{fh2005pictorial}
Pedro~F Felzenszwalb and Daniel~P Huttenlocher.
\newblock Pictorial structures for object recognition.
\newblock In {\em IJCV}, 2005.

\bibitem{girshick2015fast}
Ross Girshick.
\newblock Fast r-cnn.
\newblock In {\em ICCV}, pages 1440--1448, 2015.

\bibitem{gkioxari2014using}
Georgia Gkioxari, Bharath Hariharan, Ross Girshick, and Jitendra Malik.
\newblock Using k-poselets for detecting people and localizing their keypoints.
\newblock In {\em CVPR}, 2014.

\bibitem{gross2010multi}
Ralph Gross, Iain Matthews, Jeffrey Cohn, Takeo Kanade, and Simon Baker.
\newblock Multi-pie.
\newblock {\em Image and Vision Computing}, 28(5):807--813, 2010.

\bibitem{gui2018teaching}
Liangyan Gui, Kevin Zhang, Yu-Xiong Wang, Xiaodan Liang, Jos{\'e}~MF Moura, and
  Manuela~M Veloso.
\newblock Teaching robots to predict human motion.
\newblock In {\em IROS}, 2018.

\bibitem{he2017mask}
Kaiming He, Georgia Gkioxari, Piotr Doll{\'a}r, and Ross Girshick.
\newblock Mask r-cnn.
\newblock In {\em ICCV}, 2017.

\bibitem{iqbal2016multi}
Umar Iqbal and Juergen Gall.
\newblock Multi-person pose estimation with local joint-to-person associations.
\newblock In {\em ECCV Workshop}, 2016.

\bibitem{iqbal2018hand}
Umar Iqbal, Pavlo Molchanov, Thomas Breuel Juergen~Gall, and Jan Kautz.
\newblock Hand pose estimation via latent 2.5 d heatmap regression.
\newblock In {\em ECCV}, pages 118--134, 2018.

\bibitem{Joo_2015_ICCV}
Hanbyul Joo, Hao Liu, Lei Tan, Lin Gui, Bart Nabbe, Iain Matthews, Takeo
  Kanade, Shohei Nobuhara, and Yaser Sheikh.
\newblock Panoptic studio: A massively multiview system for social motion
  capture.
\newblock In {\em ICCV}, 2015.

\bibitem{Joo_2017_TPAMI}
Hanbyul Joo, Tomas Simon, Xulong Li, Hao Liu, Lei Tan, Lin Gui, Sean Banerjee,
  Timothy~Scott Godisart, Bart Nabbe, Iain Matthews, Takeo Kanade, Shohei
  Nobuhara, and Yaser Sheikh.
\newblock Panoptic studio: A massively multiview system for social interaction
  capture.
\newblock {\em IEEE TPAMI}, 2017.

\bibitem{joo2018total}
Hanbyul Joo, Tomas Simon, and Yaser Sheikh.
\newblock Total capture: A 3d deformation model for tracking faces, hands, and
  bodies.
\newblock In {\em CVPR}, 2018.

\bibitem{kanazawa2018end}
Angjoo Kanazawa, Michael~J Black, David~W Jacobs, and Jitendra Malik.
\newblock End-to-end recovery of human shape and pose.
\newblock In {\em CVPR}, pages 7122--7131, 2018.

\bibitem{karlinsky2012using}
Leonid Karlinsky and Shimon Ullman.
\newblock Using linking features in learning non-parametric part models.
\newblock In {\em ECCV}, 2012.

\bibitem{kazemi2014one}
Vahid Kazemi and Josephine Sullivan.
\newblock One millisecond face alignment with an ensemble of regression trees.
\newblock In {\em CVPR}, pages 1867--1874, 2014.

\bibitem{Ke_2018_ECCV}
Lipeng Ke, Ming-Ching Chang, Honggang Qi, and Siwei Lyu.
\newblock Multi-scale structure-aware network for human pose estimation.
\newblock In {\em ECCV}, 2018.

\bibitem{lan2005beyond}
Xiangyang Lan and Daniel~P Huttenlocher.
\newblock Beyond trees: Common-factor models for 2d human pose recovery.
\newblock In {\em ICCV}, 2005.

\bibitem{lin2014microsoft}
Tsung-Yi Lin, Michael Maire, Serge Belongie, James Hays, Pietro Perona, Deva
  Ramanan, Piotr Doll{\'a}r, and C~Lawrence Zitnick.
\newblock Microsoft {COCO}: common objects in context.
\newblock In {\em ECCV}, 2014.

\bibitem{mehta2017vnect}
Dushyant Mehta, Srinath Sridhar, Oleksandr Sotnychenko, Helge Rhodin, Mohammad
  Shafiei, Hans-Peter Seidel, Weipeng Xu, Dan Casas, and Christian Theobalt.
\newblock Vnect: Real-time 3d human pose estimation with a single rgb camera.
\newblock In {\em ACM TOG}, 2017.

\bibitem{misra2016cross}
Ishan Misra, Abhinav Shrivastava, Abhinav Gupta, and Martial Hebert.
\newblock Cross-stitch networks for multi-task learning.
\newblock In {\em CVPR}, pages 3994--4003, 2016.

\bibitem{mueller2018ganerated}
Franziska Mueller, Florian Bernard, Oleksandr Sotnychenko, Dushyant Mehta,
  Srinath Sridhar, Dan Casas, and Christian Theobalt.
\newblock Ganerated hands for real-time 3d hand tracking from monocular rgb.
\newblock In {\em CVPR}, pages 49--59, 2018.

\bibitem{newell2017associative}
Alejandro Newell, Zhiao Huang, and Jia Deng.
\newblock Associative embedding: End-to-end learning for joint detection and
  grouping.
\newblock In {\em NIPS}, 2017.

\bibitem{newell2016stacked}
Alejandro Newell, Kaiyu Yang, and Jia Deng.
\newblock Stacked hourglass networks for human pose estimation.
\newblock In {\em ECCV}, 2016.

\bibitem{nie2018ppn}
Xuecheng Nie, Jiashi Feng, Junliang Xing, and Shuicheng Yan.
\newblock Pose partition networks for multi-person pose estimation.
\newblock In {\em ECCV}, 2018.

\bibitem{oikonomidis2012tracking}
Iasonas Oikonomidis, Nikolaos Kyriazis, and Antonis~A Argyros.
\newblock Tracking the articulated motion of two strongly interacting hands.
\newblock In {\em CVPR}, pages 1862--1869. IEEE, 2012.

\bibitem{panteleris2017using}
Paschalis Panteleris, Iason Oikonomidis, and Antonis Argyros.
\newblock Using a single rgb frame for real time 3d hand pose estimation in the
  wild.
\newblock In {\em IEEE WACV}, pages 436--445. IEEE, 2018.

\bibitem{papandreou2018personlab}
George Papandreou, Tyler Zhu, Liang-Chieh Chen, Spyros Gidaris, Jonathan
  Tompson, and Kevin Murphy.
\newblock Personlab: Person pose estimation and instance segmentation with a
  bottom-up, part-based, geometric embedding model.
\newblock In {\em ECCV}, 2018.

\bibitem{papandreou2017towards}
George Papandreou, Tyler Zhu, Nori Kanazawa, Alexander Toshev, Jonathan
  Tompson, Chris Bregler, and Kevin Murphy.
\newblock Towards accurate multi-person pose estimation in the wild.
\newblock In {\em CVPR}, 2017.

\bibitem{frgc_2010}
P~Jonathon Phillips, Patrick~J Flynn, Todd Scruggs, Kevin~W Bowyer, Jin Chang,
  Kevin Hoffman, Joe Marques, Jaesik Min, and William Worek.
\newblock Overview of the face recognition grand challenge.
\newblock In {\em CVPR}, volume~1, pages 947--954. IEEE, 2005.

\bibitem{pishchulin2013poselet}
Leonid Pishchulin, Mykhaylo Andriluka, Peter Gehler, and Bernt Schiele.
\newblock Poselet conditioned pictorial structures.
\newblock In {\em CVPR}, 2013.

\bibitem{pishchulin2015deepcut}
Leonid Pishchulin, Eldar Insafutdinov, Siyu Tang, Bjoern Andres, Mykhaylo
  Andriluka, Peter Gehler, and Bernt Schiele.
\newblock Deepcut: Joint subset partition and labeling for multi person pose
  estimation.
\newblock In {\em CVPR}, 2016.

\bibitem{pishchulin2012articulated}
Leonid Pishchulin, Arjun Jain, Mykhaylo Andriluka, Thorsten Thorm{\"a}hlen, and
  Bernt Schiele.
\newblock Articulated people detection and pose estimation: Reshaping the
  future.
\newblock In {\em CVPR}, 2012.

\bibitem{qian2017pose}
Xuelin Qian, Yanwei Fu, Tao Xiang, Wenxuan Wang, Jie Qiu, Yang Wu, Yu-Gang
  Jiang, and Xiangyang Xue.
\newblock Pose-normalized image generation for person re-identification.
\newblock In {\em ECCV}, 2018.

\bibitem{raaj2018efficient}
Yaadhav Raaj, Haroon Idrees, Gines Hidalgo, and Yaser Sheikh.
\newblock Efficient online multi-person 2d pose tracking with recurrent
  spatio-temporal affinity fields.
\newblock In {\em CVPR}, 2019.

\bibitem{ramanan2005strike}
Deva Ramanan, David~A Forsyth, and Andrew Zisserman.
\newblock Strike a {P}ose: {T}racking people by finding stylized poses.
\newblock In {\em CVPR}, 2005.

\bibitem{ranjan2019hyperface}
Rajeev Ranjan, Vishal~M Patel, and Rama Chellappa.
\newblock Hyperface: A deep multi-task learning framework for face detection,
  landmark localization, pose estimation, and gender recognition.
\newblock {\em IEEE TPAMI}, 41(1):121--135, 2019.

\bibitem{ruder2017overview}
Sebastian Ruder.
\newblock An overview of multi-task learning in deep neural networks.
\newblock {\em arXiv preprint arXiv:1706.05098}, 2017.

\bibitem{sagonas2016300}
Christos Sagonas, Epameinondas Antonakos, Georgios Tzimiropoulos, Stefanos
  Zafeiriou, and Maja Pantic.
\newblock 300 faces in-the-wild challenge: Database and results.
\newblock {\em Image and vision computing}, 47:3--18, 2016.

\bibitem{saragih2009face}
Jason~M Saragih, Simon Lucey, and Jeffrey~F Cohn.
\newblock Face alignment through subspace constrained mean-shifts.
\newblock In {\em ICCV}, pages 1034--1041. IEEE, 2009.

\bibitem{sharp2015accurate}
Toby Sharp, Cem Keskin, Duncan Robertson, Jonathan Taylor, Jamie Shotton, David
  Kim, Christoph Rhemann, Ido Leichter, Alon Vinnikov, Yichen Wei, et~al.
\newblock Accurate, robust, and flexible real-time hand tracking.
\newblock In {\em ACM CHI}, pages 3633--3642. ACM, 2015.

\bibitem{sigal2006measure}
Leonid Sigal and Michael~J Black.
\newblock Measure locally, reason globally: Occlusion-sensitive articulated
  pose estimation.
\newblock In {\em CVPR}, 2006.

\bibitem{simon2017hand}
Tomas Simon, Hanbyul Joo, Iain Matthews, and Yaser Sheikh.
\newblock Hand keypoint detection in single images using multiview
  bootstrapping.
\newblock In {\em CVPR}, 2017.

\bibitem{simonyan2014very}
Karen Simonyan and Andrew Zisserman.
\newblock Very deep convolutional networks for large-scale image recognition.
\newblock {\em arXiv preprint arXiv:1409.1556}, 2014.

\bibitem{sridhar2015fast}
Srinath Sridhar, Franziska Mueller, Antti Oulasvirta, and Christian Theobalt.
\newblock Fast and robust hand tracking using detection-guided optimization.
\newblock In {\em CVPR}, pages 3213--3221, 2015.

\bibitem{sridhar2013interactive}
Srinath Sridhar, Antti Oulasvirta, and Christian Theobalt.
\newblock Interactive markerless articulated hand motion tracking using rgb and
  depth data.
\newblock In {\em ICCV}, pages 2456--2463, 2013.

\bibitem{tang2018deeply}
Wei Tang, Pei Yu, and Ying Wu.
\newblock Deeply learned compositional models for human pose estimation.
\newblock In {\em ECCV}, 2018.

\bibitem{wang2008multiple}
Yang Wang and Greg Mori.
\newblock Multiple tree models for occlusion and spatial constraints in human
  pose estimation.
\newblock In {\em ECCV}, 2008.

\bibitem{wei2016cpm}
Shih-En Wei, Varun Ramakrishna, Takeo Kanade, and Yaser Sheikh.
\newblock Convolutional {P}ose {M}achines.
\newblock In {\em CVPR}, 2016.

\bibitem{west1996introduction}
Douglas~B. West et~al.
\newblock {\em Introduction to graph theory}, volume~2.
\newblock Prentice hall Upper Saddle River, NJ, 1996.

\bibitem{xiang2018monocular}
Donglai Xiang, Hanbyul Joo, and Yaser Sheikh.
\newblock Monocular total capture: Posing face, body, and hands in the wild.
\newblock In {\em CVPR}, pages 10965--10974, 2019.

\bibitem{xiao2018simple}
Bin Xiao, Haiping Wu, and Yichen Wei.
\newblock Simple baselines for human pose estimation and tracking.
\newblock In {\em ECCV}, 2018.

\bibitem{xiong2013supervised}
Xuehan Xiong and Fernando De~la Torre.
\newblock Supervised descent method and its applications to face alignment.
\newblock In {\em CVPR}, pages 532--539, 2013.

\bibitem{yang2017learning}
Wei Yang, Shuang Li, Wanli Ouyang, Hongsheng Li, and Xiaogang Wang.
\newblock Learning feature pyramids for human pose estimation.
\newblock In {\em ICCV}, 2017.

\bibitem{yang2016trace}
Yongxin Yang and Timothy~M Hospedales.
\newblock Trace norm regularised deep multi-task learning.
\newblock {\em arXiv preprint arXiv:1606.04038}, 2016.

\bibitem{yang2013articulated}
Yi Yang and Deva Ramanan.
\newblock Articulated human detection with flexible mixtures of parts.
\newblock In {\em IEEE TPAMI}, 2013.

\bibitem{zhang2016joint}
Kaipeng Zhang, Zhanpeng Zhang, Zhifeng Li, and Yu Qiao.
\newblock Joint face detection and alignment using multitask cascaded
  convolutional networks.
\newblock {\em IEEE SPL}, 23(10):1499--1503, 2016.

\bibitem{zhang2014facial}
Zhanpeng Zhang, Ping Luo, Chen~Change Loy, and Xiaoou Tang.
\newblock Facial landmark detection by deep multi-task learning.
\newblock In {\em ECCV}, pages 94--108. Springer, 2014.

\bibitem{zhang2016learning}
Zhanpeng Zhang, Ping Luo, Chen~Change Loy, and Xiaoou Tang.
\newblock Learning deep representation for face alignment with auxiliary
  attributes.
\newblock {\em IEEE TPAMI}, 38(5):918--930, 2016.

\bibitem{zhu2015face}
Shizhan Zhu, Cheng Li, Chen Change~Loy, and Xiaoou Tang.
\newblock Face alignment by coarse-to-fine shape searching.
\newblock In {\em CVPR}, pages 4998--5006, 2015.

\bibitem{zimmermann2017learning}
Christian Zimmermann and Thomas Brox.
\newblock Learning to estimate 3d hand pose from single rgb images.
\newblock In {\em ICCV}, pages 4903--4911, 2017.

\end{thebibliography}
}

\end{document}